\def\templateArxiv{arxiv}
\def\templateIEEE{IEEE}
\def\templateElsevier{elsevier}

\def\templateIEEEAccess{IEEE-access}

\def\templateType{arxiv}
\def\templateSubType{IEEE-tran}

\ifx\templateType\templateArxiv
    \documentclass[lettersize, twocolumn, journal, 11pt]{IEEEtran}

\usepackage[table]{xcolor}

\usepackage[english]{babel}
\usepackage[utf8]{inputenc}

\usepackage{amssymb}
\usepackage{multicol}
\usepackage{afterpage}

\usepackage{wrapfig}

\usepackage[table]{xcolor}
\usepackage{multirow}
\usepackage{makecell}
\usepackage{array}
\usepackage{rotating}
\usepackage{booktabs}
\usepackage[skip=3pt]{caption}
\usepackage{stfloats} 

\usepackage{cellspace} 
\usepackage{subcaption}
\usepackage{setspace}
\usepackage{placeins}

\usepackage{paralist}

\usepackage{amssymb}
\usepackage{amsmath}
\usepackage{bm}

\usepackage{expl3}

\usepackage{dirtytalk}
\usepackage{gensymb}

\usepackage{hyperref}
\hypersetup{
    colorlinks,
    linkcolor={red!50!black},
    citecolor={blue!60!black},
    urlcolor={blue!70!black}
}

\usepackage{soul}
\usepackage[final]{pdfcomment}

 \usepackage{microtype}

\usepackage{tikz}

\newcommand\copyrighttext{\footnotesize
  This work has been submitted to the IEEE for possible publication. Copyright may be transferred without notice, after which this version may no longer be accessible.
}
\newcommand\copyrightnotice{\begin{tikzpicture}[remember picture,overlay]
\node[anchor=south,yshift=10pt] at (current page.south) {\fbox{\parbox{\dimexpr\textwidth-\fboxsep-\fboxrule\relax}{\copyrighttext}}};
\end{tikzpicture}}

 \else \ifx\templateType\templateIEEE
\if\templateSubType\templateIEEEAccess

\let\texyear\year

\documentclass[Path=./ieeeTemplates/]{ieeeaccess}

\let\ieeeaccessyear\year
\let\year\texyear

\usepackage[table]{xcolor}

\usepackage[english]{babel}
\usepackage[utf8]{inputenc}

\usepackage{amssymb}
\usepackage{multicol}
\usepackage{afterpage}

\usepackage{wrapfig}

\usepackage[table]{xcolor}
\usepackage{multirow}
\usepackage{makecell}
\usepackage{array}
\usepackage{rotating}
\usepackage{booktabs}
\usepackage[skip=3pt]{caption}
\usepackage{stfloats} 

\usepackage{cellspace} 
\usepackage{subcaption}
\usepackage{setspace}
\usepackage{placeins}

\usepackage{paralist}

\usepackage{amssymb}
\usepackage{amsmath}
\usepackage{bm}

\usepackage{expl3}

\usepackage{dirtytalk}
\usepackage{gensymb}

\usepackage{hyperref}
\hypersetup{
    colorlinks,
    linkcolor={red!50!black},
    citecolor={blue!60!black},
    urlcolor={blue!70!black}
}

\usepackage{soul}
\usepackage[final]{pdfcomment}

\let\year\ieeeaccessyear
\definecolor{accessblue}{cmyk}{1, 0.3, 0, 0.2}
\definecolor{greycolor}{cmyk}{0,0,0,.8}

\graphicspath{{./ieeeTemplates}{./pics/authors}}     \else
    \documentclass[journal]{IEEEtran}

\usepackage[english]{babel}
\usepackage[utf8]{inputenc}

\usepackage{amssymb}
\usepackage{multicol}
\usepackage{afterpage}

\usepackage{wrapfig}

\usepackage[table]{xcolor}
\usepackage{multirow}
\usepackage{makecell}
\usepackage{array}
\usepackage{rotating}
\usepackage{booktabs}
\usepackage[skip=3pt]{caption}
\usepackage{stfloats} 

\usepackage{cellspace} 
\usepackage{subcaption}
\usepackage{setspace}
\usepackage{placeins}

\usepackage{paralist}

\usepackage{amssymb}
\usepackage{amsmath}
\usepackage{bm}

\usepackage{expl3}

\usepackage{dirtytalk}
\usepackage{gensymb}

\usepackage{hyperref}
\hypersetup{
    colorlinks,
    linkcolor={red!50!black},
    citecolor={blue!60!black},
    urlcolor={blue!70!black}
}

\usepackage{soul}
\usepackage[final]{pdfcomment}

\graphicspath{{./ieeeTemplates}{./pics/authors}}     \fi
\else \ifx\templateType\templateElsevier

\documentclass[authoryear, preprint, review, 12pt, times, 3p]{elsarticle}

\biboptions{sort&compress}

\usepackage[english]{babel}
\usepackage[utf8]{inputenc}

\usepackage{amssymb}
\usepackage{multicol}
\usepackage{afterpage}

\usepackage{wrapfig}

\usepackage[table]{xcolor}
\usepackage{multirow}
\usepackage{makecell}
\usepackage{array}
\usepackage{rotating}
\usepackage{booktabs}
\usepackage[skip=3pt]{caption}
\usepackage{stfloats} 

\usepackage{cellspace} 
\usepackage{subcaption}
\usepackage{setspace}
\usepackage{placeins}

\usepackage{paralist}

\usepackage{amssymb}
\usepackage{amsmath}
\usepackage{bm}

\usepackage{expl3}

\usepackage{dirtytalk}
\usepackage{gensymb}

\usepackage{hyperref}
\hypersetup{
    colorlinks,
    linkcolor={red!50!black},
    citecolor={blue!60!black},
    urlcolor={blue!70!black}
}

\usepackage{soul}
\usepackage[final]{pdfcomment}

  \fi\fi\fi

\graphicspath{{.}{./images}{./images/TvS}{./images/demo}{./images/pipeline}{./images/results}
{./images/results/genByDetector}
{./images/results/graphRates}
{./images/results/solo}
{./images/process}
{./images/exDist}
}

\begin{document}

\ifx\templateType\templateArxiv
    
\title{
Evaluating Anomaly Detectors for Simulated Highly Imbalanced Industrial Classification Problems
}

\author{\IEEEauthorblockN{Lesley Wheat\IEEEauthorrefmark{1}\IEEEauthorrefmark{2}\IEEEauthorrefmark{3}, Martin v. Mohrenschildt\IEEEauthorrefmark{1} and Saeid Habibi\IEEEauthorrefmark{2}}

\IEEEauthorblockA{\IEEEauthorrefmark{1}Department of Computing and Software, McMaster University, Hamilton, Canada}

\IEEEauthorblockA{\IEEEauthorrefmark{2}Center for Mechatronics and Hybrid Technologies, McMaster University, Hamilton, Canada}

\IEEEauthorblockA{\IEEEauthorrefmark{3}Email: wheatd@mcmaster.ca}

}

\twocolumn[
\begin{@twocolumnfalse}
\maketitle
\end{@twocolumnfalse}
\copyrightnotice

\begin{abstract}
~{Machine learning offers potential solutions to current issues in industrial systems in areas such as quality control and predictive maintenance, but also faces unique barriers in industrial applications.
An ongoing challenge is extreme class imbalance, primarily due to the limited availability of faulty data during training.
This paper presents a comprehensive evaluation of anomaly detection algorithms using a problem-agnostic simulated dataset that reflects real-world engineering constraints.
Using a synthetic dataset with a hyper-spherical based anomaly distribution in 2D and 10D, we benchmark 14 detectors across training datasets with anomaly rates between 0.05\% and 20\% and training sizes between 1 000 and 10 000 (with a testing dataset size of 40 000) to assess performance and generalization error.
Our findings reveal that the best detector is highly dependant on the total number of faulty examples in the training dataset, with additional healthy examples offering insignificant benefits in most cases.
With less than 20 faulty examples, unsupervised methods (kNN/LOF) dominate; but around 30-50 faulty examples, semi-supervised (XGBOD) and supervised (SVM/CatBoost) detectors, we see large performance increases.
While semi-supervised methods do not show significant benefits with only two features, the improvements are evident at ten features.
The study highlights the performance drop on generalization of anomaly detection methods on smaller datasets, and provides practical insights for deploying anomaly detection in industrial environments.

 }
\end{abstract}


\bigskip
]

 \else \ifx\templateType\templateIEEE
\if\templateSubType\templateIEEEAccess

\history{Date of submission July 11, 2024.}
\doi{10.1109/ACCESS.2024.DOI}

\title{
Evaluating Anomaly Detectors for Simulated Highly Imbalanced Industrial Classification Problems
}

\author{\uppercase{Lesley Wheat}\authorrefmark{1}\authorrefmark{2},
\uppercase{Martin v. Mohrenschildt} \authorrefmark{1}, \uppercase{ and Saeid Habibi} \authorrefmark{2} \IEEEmembership{Member, IEEE}}
\address[1]{Department of Computing and Software, McMaster University, Hamilton, ON L8S4L7 Canada}
\address[2]{Center for Mechatronics and Hybrid Technologies (CMHT), Department of Mechanical Engineering, McMaster University}
\tfootnote{
This work was supported in part by the Canada Research Chairs (CRC) Program under Project CRC-2020-0127 and in part by the Natural Sciences and Engineering Research Council of Canada (NSERC) Ford-Mitacs Alliance under Project ALLRP-590906-23.
 }

\markboth
{Wheat \headeretal: Evaluating Anomaly Detectors for Simulated Highly Imbalanced Industrial Classification Problems}
{Wheat \headeretal: Evaluating Anomaly Detectors for Simulated Highly Imbalanced Industrial Classification Problems}

\corresp{Corresponding author: Lesley Wheat (e-mail: wheatd@mcmaster.ca).}

\begin{abstract}
Machine learning offers potential solutions to current issues in industrial systems in areas such as quality control and predictive maintenance, but also faces unique barriers in industrial applications.
An ongoing challenge is extreme class imbalance, primarily due to the limited availability of faulty data during training.
This paper presents a comprehensive evaluation of anomaly detection algorithms using a problem-agnostic simulated dataset that reflects real-world engineering constraints.
Using a synthetic dataset with a hyper-spherical based anomaly distribution in 2D and 10D, we benchmark 14 detectors across training datasets with anomaly rates between 0.05\% and 20\% and training sizes between 1 000 and 10 000 (with a testing dataset size of 40 000) to assess performance and generalization error.
Our findings reveal that the best detector is highly dependant on the total number of faulty examples in the training dataset, with additional healthy examples offering insignificant benefits in most cases.
With less than 20 faulty examples, unsupervised methods (kNN/LOF) dominate; but around 30-50 faulty examples, semi-supervised (XGBOD) and supervised (SVM/CatBoost) detectors, we see large performance increases.
While semi-supervised methods do not show significant benefits with only two features, the improvements are evident at ten features.
The study highlights the performance drop on generalization of anomaly detection methods on smaller datasets, and provides practical insights for deploying anomaly detection in industrial environments.

 \end{abstract}


\titlepgskip=-15pt

\maketitle     \else
    \title{
Evaluating Anomaly Detectors for Simulated Highly Imbalanced Industrial Classification Problems
}

\author{Lesley~Wheat,~\IEEEmembership{Graduate~Member,~IEEE,}
        Martin~v.~Mohrenschildt, and~Saeid~Habibi,~\IEEEmembership{Member,~IEEE}\thanks{
This work was supported in part by the Canada Research Chairs (CRC) Program under Project CRC-2020-0127 and in part by the Natural Sciences and Engineering Research Council of Canada (NSERC) Ford-Mitacs Alliance under Project ALLRP-590906-23.
}\thanks{
The authors are with McMaster University, Hamilton, ON L8S4L7 Canada (email: wheatd@mcmaster.ca).
}}

\markboth{
}{Wheat \MakeLowercase{\textit{et al.}}: Evaluating Anomaly Detectors for Simulated Highly Imbalanced Industrial Classification Problems}

\maketitle

\begin{abstract}
Machine learning offers potential solutions to current issues in industrial systems in areas such as quality control and predictive maintenance, but also faces unique barriers in industrial applications.
An ongoing challenge is extreme class imbalance, primarily due to the limited availability of faulty data during training.
This paper presents a comprehensive evaluation of anomaly detection algorithms using a problem-agnostic simulated dataset that reflects real-world engineering constraints.
Using a synthetic dataset with a hyper-spherical based anomaly distribution in 2D and 10D, we benchmark 14 detectors across training datasets with anomaly rates between 0.05\% and 20\% and training sizes between 1 000 and 10 000 (with a testing dataset size of 40 000) to assess performance and generalization error.
Our findings reveal that the best detector is highly dependant on the total number of faulty examples in the training dataset, with additional healthy examples offering insignificant benefits in most cases.
With less than 20 faulty examples, unsupervised methods (kNN/LOF) dominate; but around 30-50 faulty examples, semi-supervised (XGBOD) and supervised (SVM/CatBoost) detectors, we see large performance increases.
While semi-supervised methods do not show significant benefits with only two features, the improvements are evident at ten features.
The study highlights the performance drop on generalization of anomaly detection methods on smaller datasets, and provides practical insights for deploying anomaly detection in industrial environments.

 \end{abstract}


\IEEEpeerreviewmaketitle     \fi
\else \ifx\templateType\templateElsevier
    \begin{frontmatter}

\title{Paper 4}

\affiliation[inst1]{organization={Department of Computing and Software, McMaster University},addressline={1280 Main St W}, 
            city={Hamilton},
            postcode={L8S4L7}, 
            state={ON},
            country={Canada}}

\affiliation[inst2]{organization={Center for Mechatronics and Hybrid Technologies (CMHT),
Department of Mechanical Engineering, 
McMaster University},addressline={1280 Main St W}, 
            city={Hamilton},
            postcode={L8S4L7}, 
            state={ON},
            country={Canada}}

\author[inst1,inst2]{Lesley Wheat}
\author[inst1]{Martin v. Mohrenschildt}
\author[inst2]{Saeid Habibi}

\begin{abstract}

\noindent
Abstract

\end{abstract}

\begin{highlights}
\item A
\item B
\item C
\end{highlights}


\end{frontmatter} \fi\fi\fi

\section{Introduction}

With recent advances in machine learning, there has been increasing interest in applying data-driven methods to ongoing industrial challenges such as fault detection, process monitoring, quality control, and more \cite{kharitonov_comparative_2022, chevtchenko_anomaly_2023, melo_data-driven_2024, liso_review_2024}.
Data-driven methods offer the potential to automate inspection points, reduce equipment failures and improve overall efficiency, all without requiring expert knowledge of the system in question.
However, industrial applications\pdfmarkupcomment[color=yellow]{ are known to}{changed} have challenges and considerations that are not present in other machine learning applications, making it difficult to directly translate the results from existing benchmarks to specific use cases \cite{han_adbench_2022}.

In industrial settings, examples of faulty or defective conditions are often extremely limited due to multiple factors.
For example, the anomaly rate in real-world mass manufacturing problems can easily fall under 1\% \cite{risdal_bosch_2016}, which creates a significant barrier to effectively training machine learning models.
While classifiers often struggle in these scenarios, anomaly detection techniques have been formulated for the purpose of tackling these very sorts of extreme imbalances, and offer promising solutions based on existing benchmarks \cite{han_adbench_2022}.

Since real-world industrial datasets available for research are extremely scarce, method evaluation is often conducted using computer or laboratory simulated datasets \cite{mauthe_creation_2021}.
For this experiment, an existing synthetic dataset based on existing problems, featuring a non-linear problem and hyper-spherical anomaly distribution, is used to evaluate the detectors \cite{wheat_testing_2025}.
Since the probability distributions are known, an ideal anomaly detector can be created for comparison purposes.

By varying the simulation parameters to create cases where 100\% classification is not possible and generating very large testing datasets, the models can be evaluated for their ability to generalize to new data from the same problem.
Two distributions were selected, the first with two features and second with ten, which respectively represent easy and difficult cases \cite{wheat_testing_2025}.

Based on considerations specific to industrial problems, specific requirements are formulated for the detectors (Section \ref{sec:req}), which are used to create the evaluation criteria, the testing scenarios, and select most promising \pdfmarkupcomment[color=yellow]{prevailing}{changed} methods from existing research for testing \cite{han_adbench_2022}.
Detectors are evaluated for their overall performance using the area under the receiver-operator curve (AUCROC), false negative rate (FNR) and false positive rate (FPR).
Their ability to generalize is measured through the differences in performance metrics between the validation and testing sets.
The risk of selecting a poor performing detector based on validation performance is examined.
Based on these results, recommendations are made for applications and further research.

To outline, the contributions of this paper are:

\noindent
\begin{minipage}{0.95\linewidth}
\ifx\templateType\templateArxiv\smallskip\fi
\smallskip
\begin{itemize}
    \item A comparison of anomaly detection algorithms against themselves and an ideal model, under very imbalanced training conditions.
    \item Evaluation of the best unsupervised, semi-supervised, and supervised anomaly detectors given different numbers of faulty and healthy examples.
    \item Investigation of the generalizability of models, based on differences between validation and testing metrics.
\end{itemize}
\smallskip
\end{minipage}

\subsection{Industrial Application-Specific Considerations} \label{sec:challenges}

Based on consultation and literature review, a variety of ongoing challenges in industry have been selected to formulate the basis of the testing scenarios, as they may benefit from anomaly detection techniques.
These issues represent specific barriers to machine learning implementation for quality and maintenance purposes in industrial settings.
This section takes a closer look at the issues, their possible causes and how they contribute to the requirements.

\ifx\templateType\templateIEEE
\begin{figure}[t]
\centering
  \begin{minipage}{0.98\linewidth}
\centering
    \subcaptionbox{Example One: classes following Gaussian distributions \cite{wheat_testing_2025}.}
      {\includegraphics[width=0.45\linewidth]{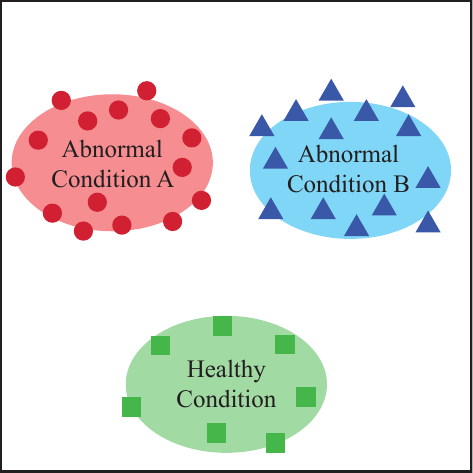}}
      \hfill
    \subcaptionbox{Example Two: difficult to model classes. Can be caused by changes in energy within vibration data \cite{wheat_testing_2025}.}
      {\includegraphics[width=0.45\linewidth]{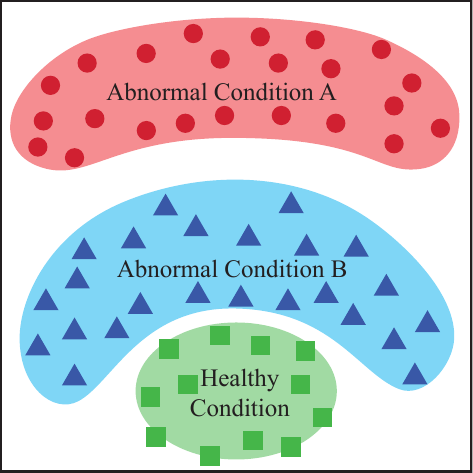}}
      \hfill
    \subcaptionbox{Example Three: top two components extracted from principal component analysis applied to CNC milling data \cite{tnani_smart_2022}. }
      {\includegraphics[width=0.85\linewidth]{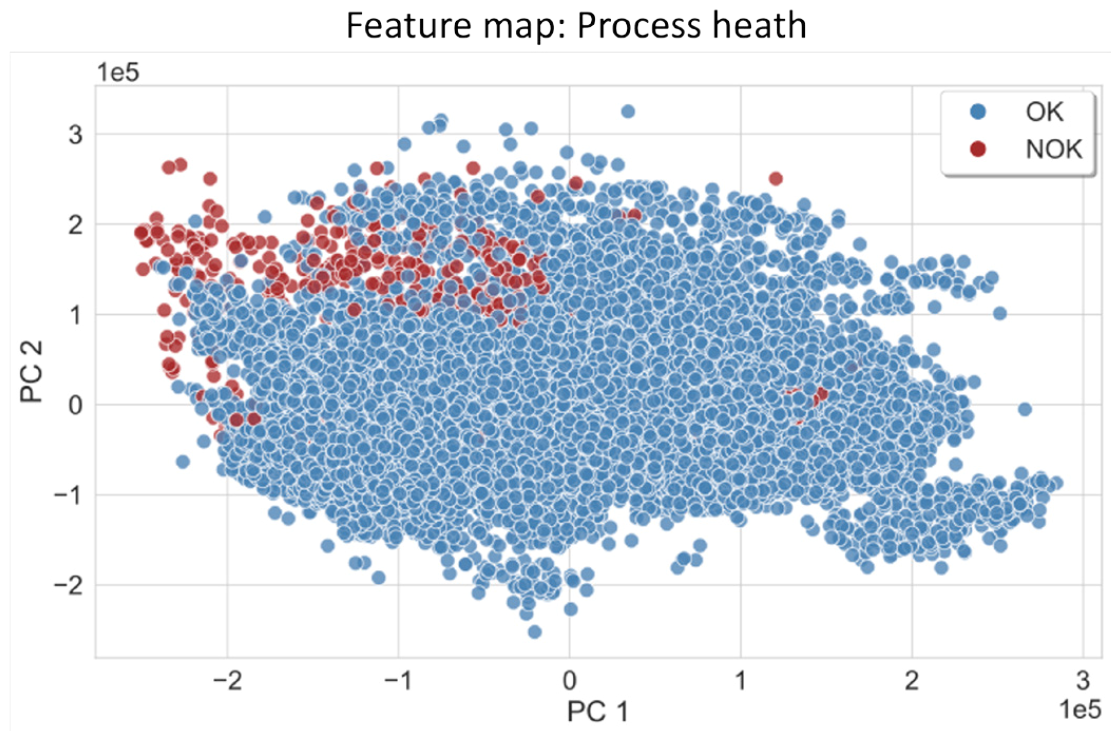}}
      \hfill
    \caption{Examples cases of structures seen in research.}
    \label{fig:ex_dist_fdd}
  \end{minipage}\quad
\end{figure}
\else
\begin{figure}[ht]
\centering
  \begin{minipage}{0.98\linewidth}
\subcaptionbox{Example One: classes following Gaussian distributions, easy to model \cite{wheat_testing_2025}.}
      {\includegraphics[width=0.48\linewidth]{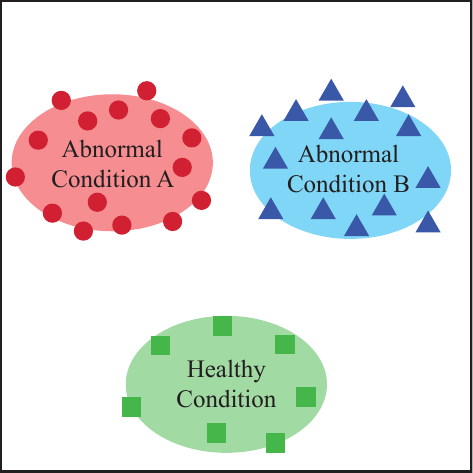}}
      \hfill
    \subcaptionbox{Example Two: difficult to model classes. Can be caused by changes in energy within vibration data (frequency domain) \cite{wheat_testing_2025}. }
      {\includegraphics[width=0.48\linewidth]{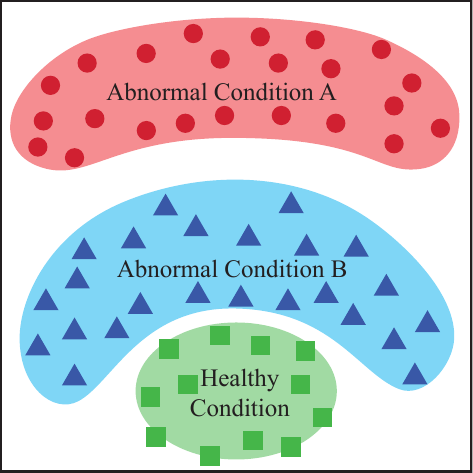}}
      \hfill
    \subcaptionbox{Example Three: top two components extracted from principal component analysis applied to CNC milling data \cite{tnani_smart_2022}. }
      {\includegraphics[width=0.98\linewidth]{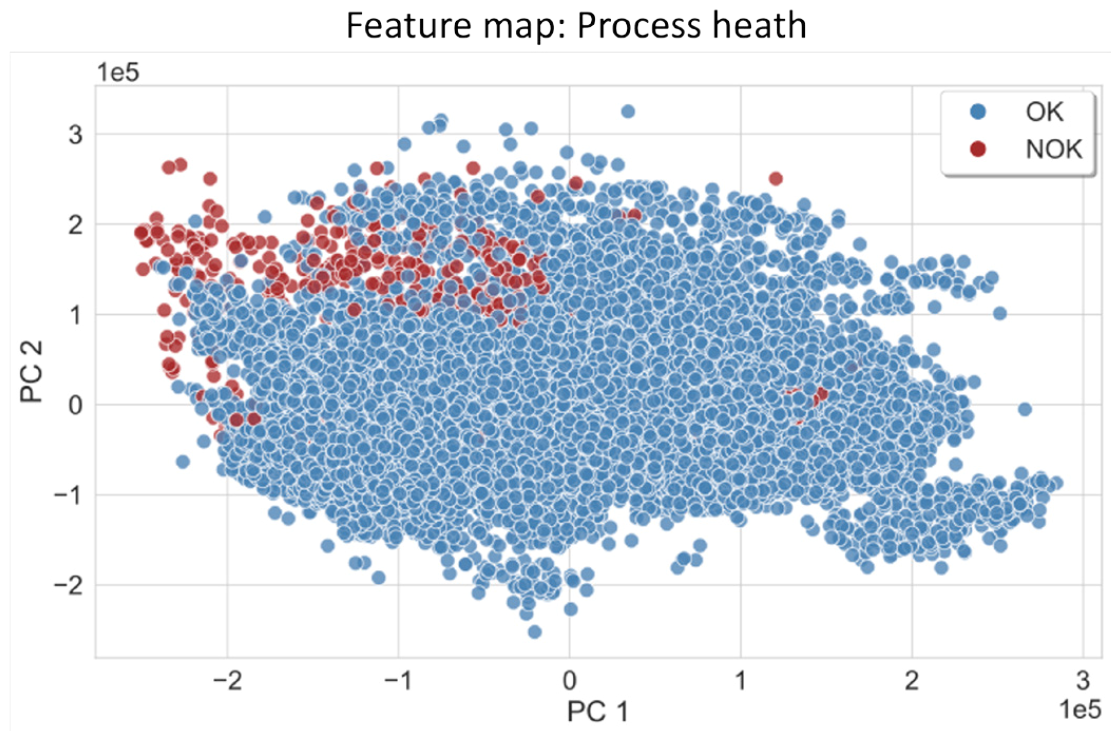}}
      \hfill
    \caption{Examples cases of structures seen in research.}
    \label{fig:ex_dist_fdd}
  \end{minipage}\quad
  \ifx\templateType\templateArxiv \vspace{-4mm} \fi
\end{figure}
\fi 
\subsubsection{Class Imbalance}

For many industrial problems, high class imbalance (under 1\%) is a significant issue \cite{jourdan_machine_2021}, meaning that a very low number of faulty or defective examples may be available for training.
Several factors contribute to this issue, including: naturally rarely occurring defects, difficulty/cost in data labelling, and the possibility of novel/unknown anomalies occurring under real-world conditions.
While some of these issues may be mitigated by how the system is implemented, or data is collected; in other cases, high class imbalance can be easily corrected.

For example, on a production line, product quality may already be very high at the point that the detection system is to be introduced \cite{tnani_smart_2022}.
Depending on the system and kind of defect to be detected, there may already be existing quality checks that would reduce the number of defective products that make it to the data collection point.
This is a particular issue for end-of-line (EOL) testing, as this should be the point in the system with the highest quality levels, thus making it one of the hardest points for collecting data on defects.

Ideally, the data collection would occur on the line as close as possible to where the defects are introduced.
However, this is not always possible, or practical.
For example, some tests may only work on a completed, or partially completed product.
Or, it may not be known at what stage of the production process a defect is introduced.

Additionally, the resources required to perform the tests needed to gather data labels may be prohibitive.
Some tests may require skilled labour, specialized equipment, and/or be destructive in nature.
By nature, destructive quality tests can only be performed on a small number of products, and are the only option for some applications.
These obstacles can make it much more difficult to collect labelled data from a production system.

Another option would be to purposely introduce defects or damage \pdfmarkupcomment[color=yellow]{product(s)}{possible plural} for the purpose of collecting data.
Obviously, intentionally damaging \pdfmarkupcomment[color=yellow]{goods}{changed} is not a preferable option, and cost and waste is not the only issue with this approach.
It is vital that the data from the artificial defects match what would be expected from natural causes.
Unfortunately, some defects can be difficult to replicate \cite{mauthe_creation_2021}.
While artificial data may still offer benefit for training, as long as the distributions are different, it does not fully substitute real data.

Also, to create a defect for the purposes of data collection, it is assumed that the type (or types) of defect being searched is known.
This is a different problem than simply attempting to detect \textbf{any} defect, including ones that may not have been previously seen.
These are two different approaches, of which the choice depends on other factors: such as, if the system needs to identify if the product has a defect or what the defect is (diagnosis).
In this case, we'll assume that we do not have examples of every defect that we want the system to detect.
Given that these are cases where such a small number of faulty examples are available, this is a reasonable assumption.

\subsubsection{Prior Knowledge}

In this paper, \say{anomaly} data is typically referenced as a single class, however, many different types of defects may be labeled as anomalous.
So, while an individual defect may follow a Gaussian distribution (a combination of its own distribution plus measurement error); as the number of possible defects increases, it becomes less likely that the distribution of the overall anomaly class is Gaussian.
Some defects, either naturally or due to feature extraction techniques, follow distributions that are difficult to model, as shown in Figure \ref{fig:ex_dist_fdd}.B.

In such cases, \say{model-free} or \say{data-driven} methods, where the underlying physical model of the problem is not known, and where a more general model is tuned using the data itself, are preferable approaches.
Generalization issues may occur because the learned model is based on the data, not knowledge of the system; and may over-fit to that data, resulting in unexpected deviations in performance on the system, when deployed in the real world.
Identifying and correcting for generalization issues is still an ongoing research problem, but it is key that the models perform accurately and consistently on the problems they are trained on.

\ifx\templateType\templateIEEE
\begin{figure}[htb]
\centering
\begin{minipage}{0.47\linewidth}
    \centering
     \includegraphics[width=1\textwidth]{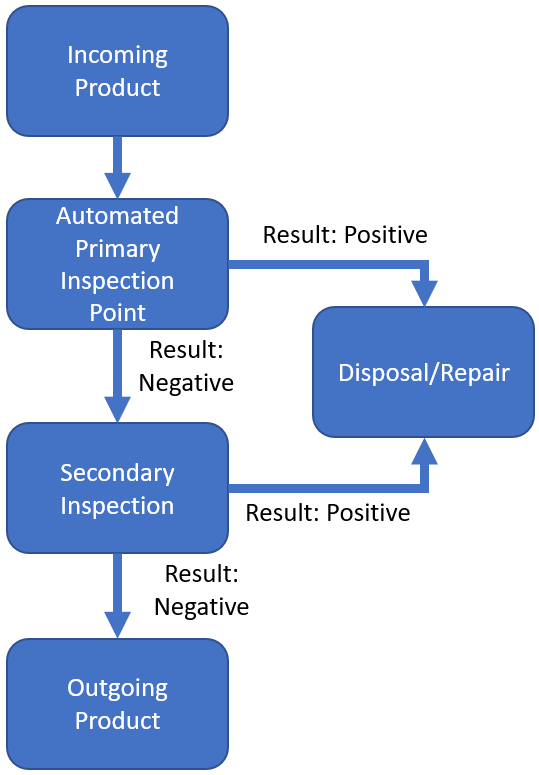}
    \caption{Example where false positives may be prioritized as they result in direct waste, whereas false negatives can be caught by a secondary inspection.
    }
    \label{fig:proccess_LFPR}
\end{minipage}\quad
\hfill
\begin{minipage}{0.47\linewidth}
\centering
     \includegraphics[width=1\textwidth]{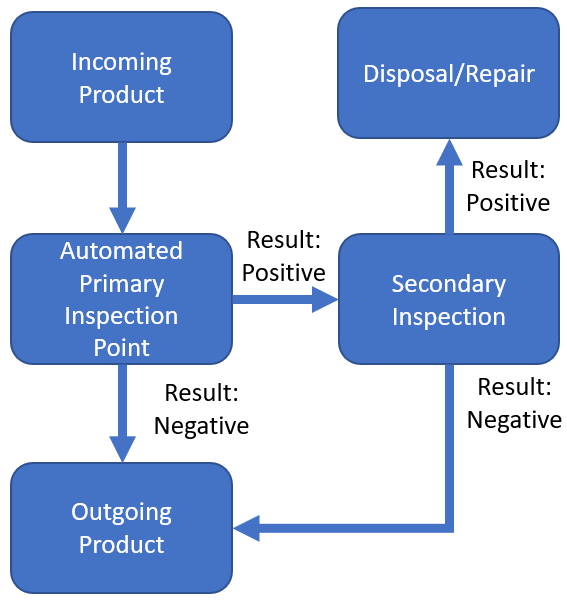}
    \caption{Example where false negatives may be prioritized, as product directly exits system after passing screening.}
    \label{fig:proccess_LFNR}
\end{minipage}
\end{figure}
\else
\begin{figure}[htb]
    \centering
     \includegraphics[width=0.3\textwidth]{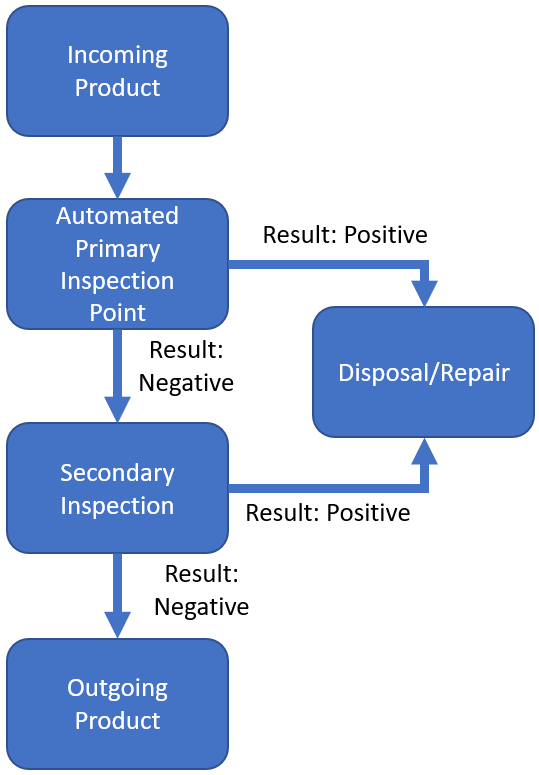}
    \caption{Example where false positives may be prioritized as they result in direct waste, whereas false negatives can be caught by a secondary inspection.
    }
    \label{fig:proccess_LFPR}
\end{figure}

\begin{figure}[htb]
    \centering
     \includegraphics[width=0.3\textwidth]{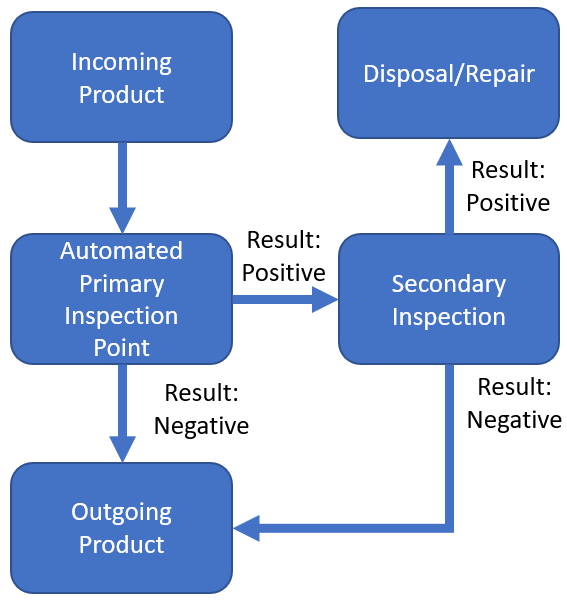}
    \caption{Example where false negatives may be prioritized, as product directly exits system after passing screening.}
    \label{fig:proccess_LFNR}
\end{figure}
\fi 
\subsubsection{Ability to Prioritize}
Depending on how a particular quality system is integrated into the production line, the False Positive Rate (FPR) or False Negative Rate (FNR) may be more important than the overall error rate.
A false negative on a defective product means it moves onto the next stage, an undesirable result, but the severity of the detection failure is fully dependent on the application.
In the case of testing food for dangerous bacteria, a false negative could lead to consumer hospitalization.
However, in the case of failing to detect a cosmetic issue, such as a paint bubble, the risk of customer dissatisfaction is much lower.
On the opposite side, a high false positive rate may lead to needless waste, expensive downtime, or alarm fatigue (loss of trust in the detection system).
Depending on the consequences and severity of false negatives and false positives, different applications may prioritize the cases differently.

On top of that, these quality systems exist as part of a larger production system which may contain multiple testing points.
Figures  \ref{fig:proccess_LFPR} and \ref{fig:proccess_LFNR} show examples of two different setups where a single automated quality system may be combined with a second, more accurate quality check.
While this does require the use of two systems rather than one, an obvious downside, it can still be useful.
For example, the secondary inspection may be worker-operated and the first inspection may be automated, so that the automated system acts as a rapid screening tool and reduces the overall workload.

In any case, the flexibility of being able to prioritize the correct detection of positives over negatives, or vice versa, is a key feature in allowing the detectors to be adapted into multiple use cases.
To go even further, the output of an anomaly \say{score} allows for even more flexibility for applications and decision-making.
Because of this, we are going to focus on detection methods that output scores rather than labels.

\subsubsection{Summary} \label{sec:req}

Based on these considerations, the anomaly detection technique should have the following characteristics:

\noindent
\begin{minipage}{0.95\linewidth}
\smallskip
\begin{enumerate}[{R}.1]
\item Require very low numbers of faulty examples. \label{req:imbalance}
\item Produce generalizable results on the problem. \label{req:generalize}
\item Produce consistent results on the problem. \label{req:consistent}
\item Do not require prior information about the problem (data-driven).\label{req:dist}
\end{enumerate}
\smallskip
\end{minipage}

Outside of these requirements, there are additional preferential characteristics.
There is a preference for more \say{black box} style methods, with less reliance on low numbers of user-defined parameters.
Methods which produce deterministic results (not utilizing random seeds) are more desirable.
Given the variety of input data across different applications, consistent performance over different numbers of features is also desirable.

\renewcommand{\arraystretch}{1.2}
\begin{table*}[htb]
\centering
\ifx\templateType\templateIEEE
\small
\fi

\begin{tabular}{lllllll}

Category & Source & Name & Year & \makecell[l]{Number of\\Examples} & \makecell[l]{Anomaly\\Rate} \\\hline \hline

Product Quality & \cite{risdal_bosch_2016} & Bosch Production Line & 2016 & 1,183,747 & 0.6\% \\ \hline

Product Quality & \cite{tnani_smart_2022} & Bosch CNC Machining & 2022 & 851 & 4\%* \\ \hline

Tool Health & \cite{von_birgelen_self-organizing_2018}
& Blade Wear Sample & 2018 & 6 & 50\% \\ \hline
Tool Health & \cite{von_birgelen_self-organizing_2018} 
& Blade Wear One Year & 2018 & 519 & N/A \\ \hline

System Monitoring & \cite{noauthor_phm_2015} & Plant Fault Detection & 2015 & 672,530 & 30\% \\ \hline

System Monitoring & \cite{noauthor_versatile_nodate} & Versatile Production & 2018 & 10,529 & N/A \\ \hline

System Monitoring & \cite{noauthor_e-coating_nodate} & Tinsley E-coating & 2020 & 111,361 & N/A \\ \hline

Monitoring & \cite{martin_del_campo_barraza_dataset_2018} 
& Wind Turbines & 2018 & 40,430 & N/A \\ \hline
Monitoring & \cite{unknown_aps_2016} & APS & 2018 & 60,000 & 2\% \\ \hline
Monitoring & \cite{backblaze_hard_2025} & Hard Drives & 2025 & 452,991,106 & 1\% \\ \hline

\end{tabular}

\caption{Summary of open-source production datasets  \cite{mauthe_creation_2021}.
N/A indicates unlabelled data.
\\ * True production rate is reported to be lower.
}
\label{table:litrev-prod}

\end{table*} 
\section{Background} \label{sec:background}

Previous work benchmarking anomaly detection algorithms for manufacturing has been studied for different applications, including images of products \cite{xie_im-iad_2024}, production schedules \cite{kharitonov_comparative_2022}, additive manufacturing \cite{xames_systematic_2023}, sensor data \cite{chevtchenko_anomaly_2023} and more.
This section will briefly cover the different types of data-driven algorithms and datasets used for evaluation.
Several detectors across these categories are selected for testing, based on previous research (discussed further in Section \ref{sec:detectors}).

\subsection{Anomaly Detection Strategies}

Anomaly detection algorithms are often categorized by how they use labels: unsupervised, semi-supervised and supervised.
Labels indicate if any sample is known to be healthy or faulty (assumed to be binary at this point).
Within each of these categories, further categories of algorithms exist (more detailed coverage can be found in \cite{han_adbench_2022}).

In unsupervised approaches, no labelling information is required.
Methods in this category focus on learning the distribution of the data, thus identifying new data as anomalous when it is \say{dissimilar} to the training data.
These methods are closely related to outlier detection and known to have difficulty detecting clustered anomalies (when present in training data).
The absence of class information makes determining a class boundary difficult, so some methods require the user to provide the expected number of faulty examples in the training dataset
(the {\say{contamination}} rate).

This category comprises the most classical machine learning techniques, such as density estimation (kNN and LOF) and clustering-based methods (CBLOF).
It also includes unsupervised adaptations of classification algorithms, such as IForest and OCSVM.
A few deep learning (DL) methods (a sub field of multilayer neural network architectures) have also been adapted to unsupervised anomaly detection but continue to struggle with a lack of labels \cite{han_adbench_2022}.

Supervised approaches require labels for all training data and employ a classification strategy.
In cases with a very large class imbalance, class re-weighting can be applied to prevent the classifier from outright favouring the majority class.
However, these methods are known to be prone to overfitting with small numbers of examples and/or high numbers of features.
They also may not perform well when the anomaly data present in the training dataset is not similar to the anomaly data in the test set (for example, if the fault conditions are different) \cite{han_adbench_2022}.

Lastly, semi-supervised learning is a combination of unsupervised and supervised learning strategies, which covers a very broad range of strategies; in an attempt to utilize the advantages of both approaches.
For example, the use of an unsupervised dimensionality reduction technique with a classifier would fall into this category.
For labels, from purely a black-box perspective, semi-supervised methods may require some partial labelling or complete labelling of all training data.

The two main types of semi-supervised anomaly detection algorithms are DL-based, and ensemble-based; which involve combining the results of multiple detectors to improve overall performance.
In XGBOD, the output of unsupervised detectors is combined with boosting algorithms to create a semi-supervised detector.
Similarly, DL methods have also been developed which combine unsupervised (for example, autoencoders) and supervised techniques (for example, a fully connected neural network).

Semi-supervised methods have shown promise in areas where fully supervised methods struggle \cite{han_adbench_2022}.
(A review of deep learning strategies for manufacturing by sub-application is covered in \cite{liso_review_2024}.)
While DL has recently risen in popularity, other methods still make up the majority of algorithms applied to sensor data \cite{chevtchenko_anomaly_2023}.

Unsupervised and semi-supervised are the most common types of algorithms applied to manufacturing problems.
However, research does suggest that semi-supervised methods are best, even when very low numbers of examples are available \cite{han_adbench_2022}.
There are no guidelines for selecting between unsupervised, semi-supervised or supervised strategies, based on available dataset size outside of specific problems (e.g. images defect detection).

\subsection{Available Datasets} \label{sec:lit-datasets}

Comparisons of anomaly detection algorithms primarily rely on simulated data (real-world or computer-generated) due to the  lack of open-source production datasets \cite{jourdan_machine_2021, mauthe_creation_2021}.
A summary of available production datasets is covered in Table \ref{table:litrev-prod}, giving an idea of the large number of examples and low anomaly rates that form the problem.
Due to the small number of production datasets, anomaly detectors are often benchmarked with simulated data or other benchmark datasets \cite{han_adbench_2022, jourdan_machine_2021}, which have higher anomaly rates.
However, many techniques have been applied to control the anomaly rate including:

\noindent
\begin{minipage}{0.95\linewidth}
\ifx\templateType\templateArxiv\smallskip\fi
\smallskip
\begin{itemize}
    \item Stratified Sampling: Keeping the anomaly rate in the training and testing datasets the same \cite{han_adbench_2022}.
    \item Modified Labels: Class labels may be changed to make the problem more difficult \cite{han_adbench_2022}.
    \item Downsampling: Reducing the number of examples by removing examples from the dataset \cite{campos_evaluation_2016}. By deleting healthy or anomalous examples, the anomaly rate can be controlled.
\end{itemize}
\smallskip
\end{minipage}

These methods do have different underlying effects on the experimental setup.
For example, changing some anomaly labels to healthy may reduce the apparent number of anomalous examples in the training dataset \cite{han_adbench_2022}, but that data is still present and may change the behaviour of anomaly detectors which make use of the labels.
Note that any combination of the methods listed above means that the labelled anomaly rate for any dataset may be artificially lowered to under a 1\% total anomaly rate.
Additionally, the ratios used for train-test splitting do also directly influence the amount of data available for training and the number of anomalous examples.

The low number of anomalous examples in production datasets makes it difficult to properly measure performance.
This is one reason why synthetically generated computer datasets have been used for testing detectors.
Some synthetic datasets use simulation models based on existing processes \cite{matzka_explainable_2020}, but others also use Gaussian distributions and similar clusters \cite{pei_synthetic_2006, gao_fuzzy_2025, sanchez_vinces_comparative_2025}.
In this case, a subset of the dataset from \cite{wheat_testing_2025} is used, as it represents a difficult scenario and includes known probability distributions for both faulty and healthy.

\begin{figure}[tb]
  \begin{minipage}{\linewidth}
\centering
    \ifx\templateType\templateIEEE
    \subcaptionbox{2D example.}
      {\includegraphics[width=0.4\linewidth]{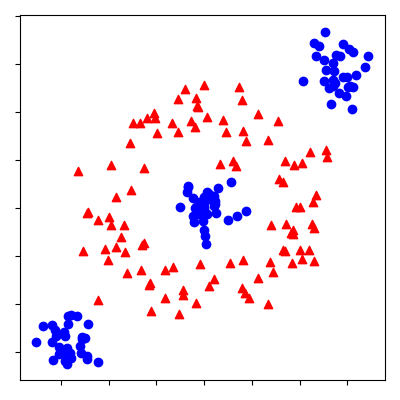}}
      \else
    \subcaptionbox{2D example with low noise.}
      {\includegraphics[width=0.4\linewidth]{2d_TvS.png}}
    \subcaptionbox{2D example with noise.}
      {\includegraphics[width=0.4\linewidth]{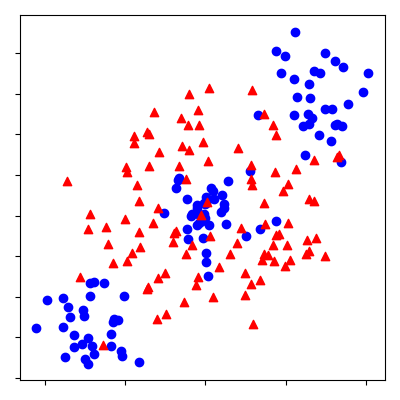}}
      \hfill
      \fi
    \subcaptionbox{3D Example.}
      {\includegraphics[width=0.45\linewidth]{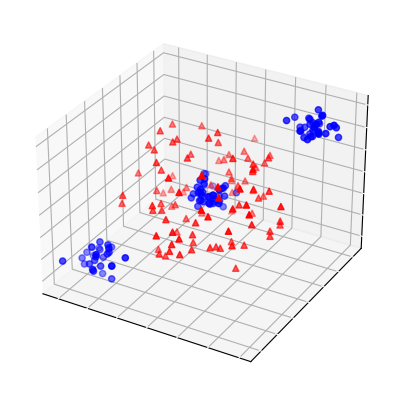}}
    \caption{Examples of TvS distribution \cite{wheat_testing_2025}.}
    \label{fig:ex-tvs}
  \end{minipage}\quad
\end{figure}

 \renewcommand{\arraystretch}{1.2}
\begin{table}[tb]
\centering
\ifx\templateType\templateIEEE
\small
\fi
\begin{tabular}{lll}

Class & \makecell[l]{Distribution Type} & Parameters \\
\hline

Healthy
& \makecell[l]{Three-mode\\Gaussian\\Mixture}
& \makecell[l]{$\sigma^2_A > 0.01$, $\mu > 0$, \\
$\mu_A = \sqrt{\mu^2/d}$,
\ifx\templateType\templateIEEE\else
\\
\fi
$\mu_1 = \vec0$, \\
$\mu_2 = \mu_A \vec1$,
\ifx\templateType\templateIEEE\else
\\
\fi
$\mu_3 = -\mu_A \vec1$} \\
\hline
Faulty
& \makecell[l]{Gaussian Mixture\\Hypersphere\\Approximation}
& \makecell[l]{ $r_B = \mu / 2$, \\
$\sigma^2_B > 0.1$} \\
\hline

\end{tabular}

\caption{Allowed parameters for TvS distribution type \cite{wheat_testing_2025}.
\ifx\templateType\templateIEEE\else
$d$ is the number of features.
\fi
}
\label{table:dparam}

\end{table} 
\subsection{Simulated Dataset} \label{sec:tvs}

This experiment will make use of the \say{TvS} distribution from \cite{wheat_testing_2025}, where the healthy data follows a spaced-out three-mode Gaussian mixture, and the faulty data forms a hypersphere (see Figure \ref{fig:ex-tvs}).
This works well as a test case because it produces a scenario where all features contribute to the class separability, the number of features can be controlled and the faulty class is spread out.

Table \ref{table:dparam} covers the distribution parameters.
$\mu$ controls the offset while $\sigma_A$ and $\sigma_B$ control the noise levels for the two classes, as shown in Figure \ref{fig:ex-tvs}.
By varying the $\mu$, $\sigma_A$ and $\sigma_B$ user-controlled parameters, the overall separability of the two classes can be controlled.

Approximately 200 clusters are used to create the hypersphere approximation, and when a small number of faulty examples are available, this makes it difficult to model the faulty distribution.
However, it is not expected that the faulty distribution would be easy to model, predictable, or consistent due to the nature of the faults.
Therefore, this is a good test case for our scenario.

This distribution does ensure a case where all features are \say{useful} and the number of features can be controlled.
Even in higher dimensions, the structure of the hypersphere still produces a similar problem, although the difficulty of the problem will naturally increase due to the curse of dimensionality.

Although the features have no contextual meaning, every feature is required in order to obtain the best classification rates.
While the Gaussian nature of the structure and small feature space does most closely match sensor data compared to other forms of input data, this could be considered as a case where the original data has already been transformed.
As the original input data may be in the form of signals, images or other types of information, there would be an assumption that proper feature extraction and dimensionality reduction has already been applied.
Thus, only the distribution and underlying separability are of concern.

A common concern in anomaly and outlier detection, is what type of anomaly is being searched for.
As this is a classification problem, the anomalies are simply examples generated by the anomaly class.
However, it means that the anomalies may not fit into typical anomaly categories such as \say{local} or \say{global}.
Thus, the results of the detectors may not be as predictable.

This dataset also supplies the probability distributions, which allows the calculation of the ideal case.
This means that the problem difficulty can be controlled, as in a problem can be created where 100\% classification is not possible.

\begin{figure}[ht]
  \begin{minipage}{\linewidth}
\centering
    \subcaptionbox{Contour plot of class probability distributions for an 2D example distribution with 50 healthy and 10 anomaly random samples. Parameters: $\mu = 1.8$, $\sigma^2_A = 0.1$ and $\sigma^2_B = 0.06$ (see Table \ref{table:dparam} for parameter explanation).}
      {\includegraphics[width=0.95\linewidth]{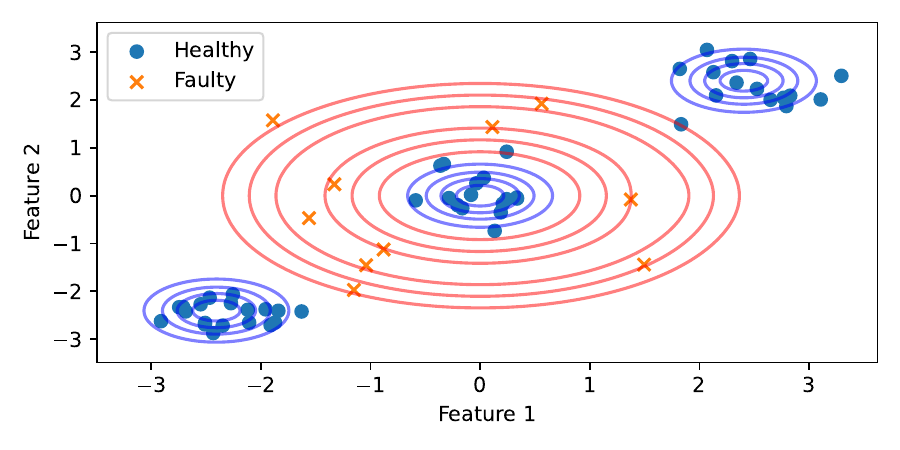}}
      \hfill
    \subcaptionbox{Kernel Density Estimate constructed based on 10000 samples of each class from distribution (a). Scores of samples shown in (a) are included below the graph.}
       {\includegraphics[width=0.95\linewidth]{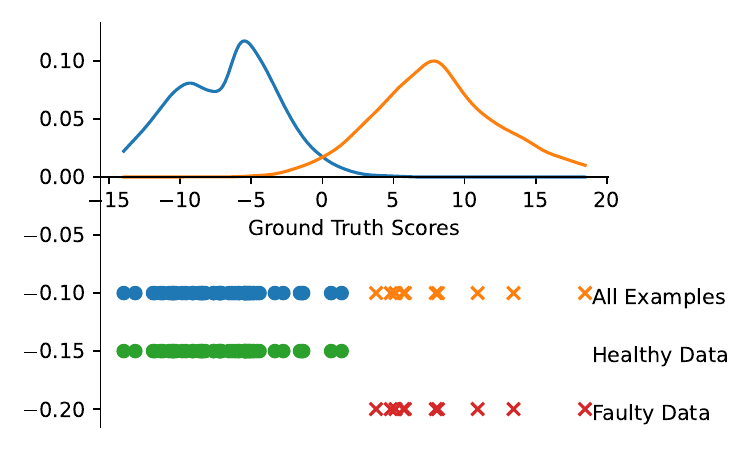}}
      \hfill
\caption{Example of class distributions and scoring distributions for one TvS scenario.}
    \label{fig:ex-gt}
  \end{minipage}\quad
\end{figure}
 \ifx\templateType\templateIEEE\else
\begin{figure}[ht]
  \begin{minipage}{\linewidth}
\centering
    \includegraphics[width=0.7\linewidth]{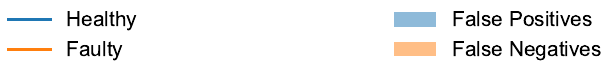}
    \subcaptionbox{False positive target rate of 5.00\% with a FNR of 1.69\%.
    }
      {\includegraphics[width=0.95\linewidth]{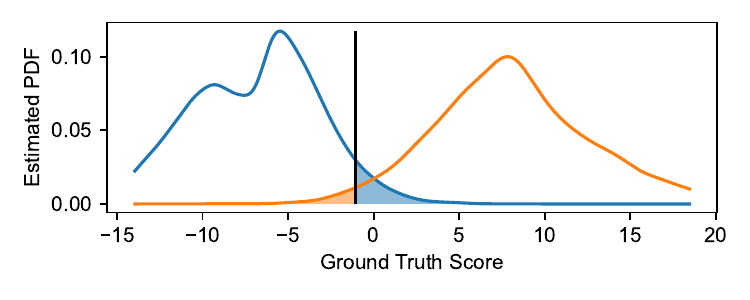}}
      \hfill
    \subcaptionbox{False negative target rate of 5.00\% with a FPR of 1.42\%.
    }
      {\includegraphics[width=0.95\linewidth]{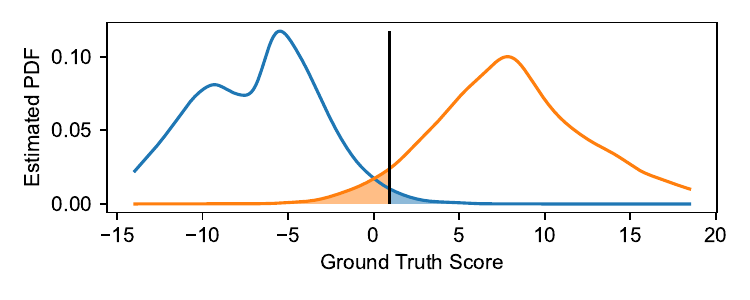}}
    \caption{Examples of target error rates for distribution in Figure \ref{fig:ex-gt}. Estimated PDF is created using kernel density estimation.}
    \label{fig:ex-gt-thres}
  \end{minipage}\quad
\end{figure}
\fi

\begin{figure}[ht]
    \centering
    \ifx\templateType\templateIEEE
    \includegraphics[width=0.95\linewidth]{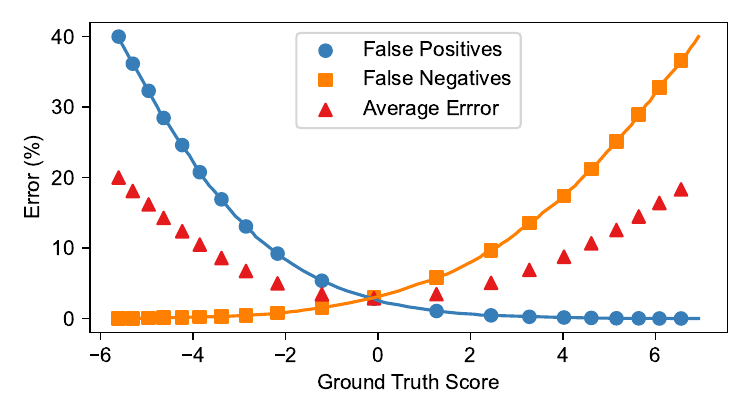}
    \else
     \includegraphics[width=0.95\linewidth]{demoError}
     \fi
    \caption{Trade off between false positives and false negatives based on different thresholds for the distribution in Figure \ref{fig:ex-gt}.}
    \label{fig:ex-fp-fn}
\end{figure} 
\section{Ground Truth Calculation} \label{sec:gt}

In order to evaluate the performance of the detectors against ideal results, an optimal or \say{ground truth}(GT) anomaly detector is required.
Here, the GT anomaly detector is constructed based on the Bayes Classifier, which can only be used when the class probability distributions are known.
Given the probability distribution function for the healthy data is $\mbox{PDF}_H$ and faulty is $\mbox{PDF}_F$, with data point $x$, the predicted label by the Bayes Classifier is  \cite{devroye_probabilistic_1996}:
\begin{equation} \label{eq:BC}
\hat{y} =
\begin{cases}
  1, & \text{if } \mbox{PDF}_F (x) > \mbox{PDF}_H (x) \\
  0, & \text{if } \mbox{PDF}_F (x) < \mbox{PDF}_H (x).
\end{cases}
\end{equation}
While this classifier has the lowest overall classification error rate on the basis of probability, it does not natively allow for control over the FPR and FNR.
In order to obtain finer control over the error rate, Eq. \ref{eq:BC} can be turned into a relative anomaly score:
\begin{equation}
g = \log \frac{\mbox{PDF}_F (x)}{\mbox{PDF}_H (x)},
\end{equation}
\begin{equation}
g = \log [\mbox{PDF}_F (x)] - \log [\mbox{PDF}_H (x)].
\end{equation}
The ratio is taken here because the probability values can get very small in higher dimensions and the conversion to logarithms can help mitigate the chance of floating point errors.
To be consistent with other anomaly detection methods, higher scores indicate a higher likelihood of an anomaly.
Figure \ref{fig:ex-gt} shows how the TvS distribution may be used to produce anomaly scores.
While this does provide us with an ideal scoring function, those values must still be converted into labels for the purpose of calculating error rates.

\subsection{Simple Anomaly Predictor} \label{sec:simple}

In order to convert the anomaly scores $g$ into a classification model to calculate error rates, a threshold must be set to decide how to classify new points.
A simple way to do this is to set the threshold based on the empirical inverse cumulative distribution function (percentile/quantile function) for one class based on the available data and target error rate.
From that threshold, the error rate from the opposite class can then be estimated from its own empirical cumulative distribution 
\ifx\templateType\templateArxiv
function (see Figure \ref{fig:ex-gt-thres}).
\else
function.
\fi
By repeating this process for a range of target percentiles for both classes, an approximate curve of the FPR-FNR trade off can be constructed (see Figure \ref{fig:ex-fp-fn}).

To calculate the estimated FNR ($\widehat{\mbox{FNR}}$) from a target FPR ($\mbox{FPR}_T$) using estimates of the CDF and ICDF from the anomaly scores of the healthy and faulty datasets:

\noindent
\begin{minipage}{0.95\linewidth}
\ifx\templateType\templateArxiv\smallskip\fi
\smallskip
\begin{equation} \label{eq:fpr_to_thres}
\hat{t} = \widehat{ICDF_H}(1-\mbox{FPR}_T),
\end{equation}
\begin{equation} \label{eq:thres_to_fnr}
\widehat{\mbox{FNR}} = \widehat{CDF_F}(\hat{t}),
\end{equation}
\noindent where $\hat{t}$ represents the estimated threshold.
\smallskip
\end{minipage}

\pdfmarkupcomment[color=yellow]{The threshold $\hat{t}$ can be used as a decision boundary to convert any scoring function into a prediction function.}{Edited}
This method also be used on the scores obtained from other anomaly detectors (see Section \ref{sec:metrics}).

\begin{figure}[tb]
\centering
     \includegraphics[width=1\linewidth,trim={0 3cm 0 0},clip]{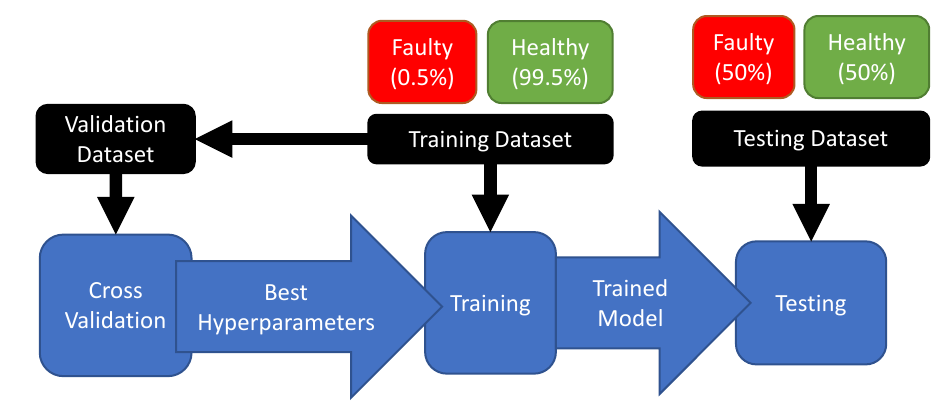}
    \caption{Model training and testing process.}
    \label{fig:pipeline}
\end{figure}

\ifx\templateType\templateArxiv
\begin{figure}[tb]
  \begin{minipage}{\linewidth}
    \centering
     \includegraphics[width=0.9\linewidth, trim={0 2.5cm 17cm 0}, clip]{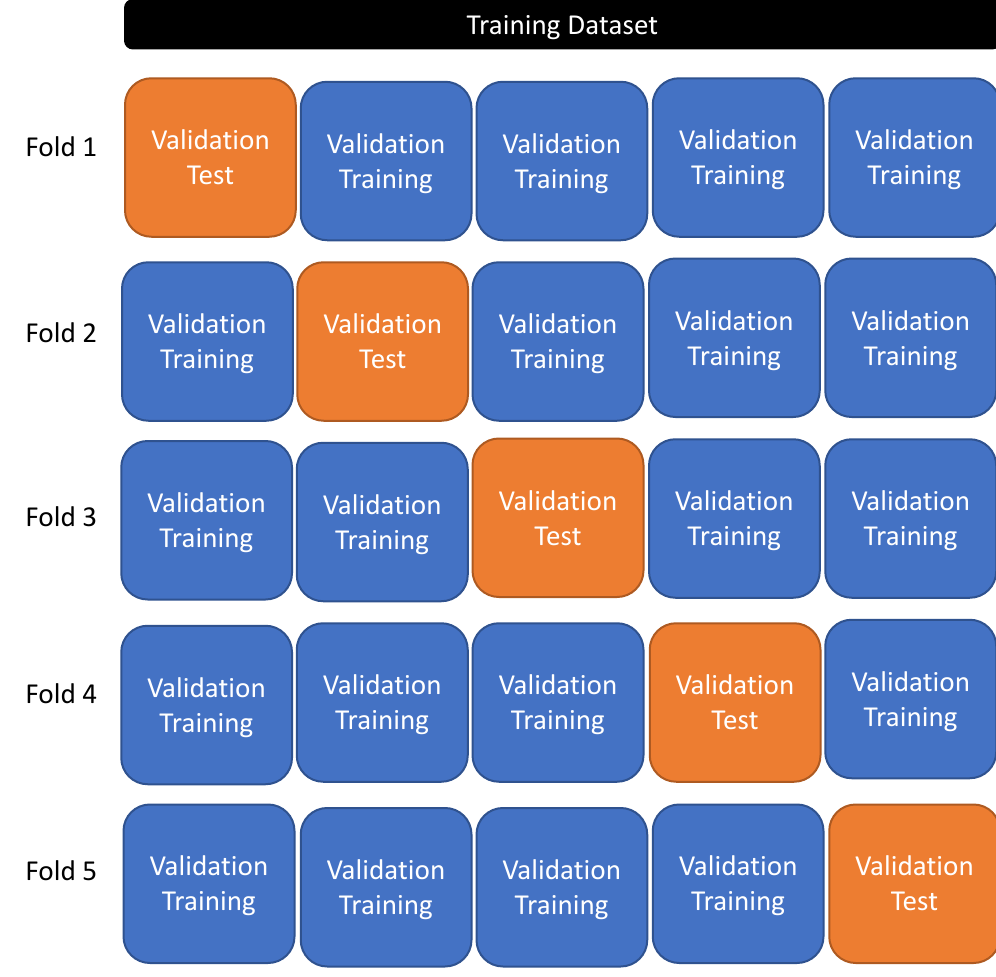}
    \caption{Visual diagram of splitting training dataset for validation.}
    \label{fig:cross-val}
  \end{minipage}\quad
\end{figure}
\fi
 \section{Experiment}

Each simulation begins by generating the training dataset based on the distribution parameters (Section \ref{sec:scenarios}) and simulation parameters (Table \ref{table:sim-param}).
The anomaly rate for the training dataset is set to 0.5\%, which represents a very high class imbalance.
Small deviations are introduced in the number of training examples (Table \ref{table:sim-param}) to introduce some variability.
A different preset seed is used for every simulation to ensure reproducibility.

Once the training dataset is created, the validation datasets are created.
These datasets are used for calculating validation metrics (Section \ref{sec:metrics}) and hyperparameter tuning (see Section \ref{sec:CV}).
Each detector has access to the same validation datasets.

Next, the detectors are retrained using the entire training dataset and applied to the testing dataset (see Figure \ref{fig:pipeline}).
The detectors are given access to all the training data (and labels) to give them the best chance of success at the problem, as would be done in deployment.
In this case, all the training data represents what is available and the testing data represents the world.

\renewcommand{\arraystretch}{1.1}
\begin{table}[tb]
\ifx\templateType\templateIEEE
\small
\fi
\centering
\begin{tabular}{ll}

Basic Simulation Parameters & Values \\
\hline

Target FPR & 1\% \\ \hline
Number of Training Samples &
\makecell[l]{$1 000\pm 2$, $2 000\pm 2$, \\$5 000\pm 2$, $10 000\pm 2$
} \\ \hline
Faulty Rate in Training Samples & 0.5\%* \\ \hline
Number of Testing Samples & $40960^\dagger$ \\ \hline
Faulty Rate in Testing Samples & 50\% \\ \hline
\makecell[l]{Number of Simulations\\per Sample Size\\per Distribution Scenario} & 100 \\ \hline

\end{tabular}

\caption{Simulation Parameters.\\
* Further examination includes rates of 0.05\% to 20\%.\\
$^\dagger$ Composed of 40 batches of 1024.
}
\label{table:sim-param}

\end{table} \renewcommand{\arraystretch}{1.1}
\begin{table}[tb]
\centering
\ifx\templateType\templateIEEE
\small
\fi
\begin{tabular}{cccc}

Scenario & \makecell[c]{Distribution
\ifx\templateType\templateIEEE\else
\\
\fi
Parameters} & \makecell[c]{Ground Truth Metrics} \\
\hline

$S_1$
& \makecell[l]{
$d = 2$, $\mu = 2.8$\\
$\sigma^2_A = 0.05$, $\sigma^2_B = 0.4$\\
} &
\makecell[l]{
$\mbox{FPR} = 1.0\%$,\\
$\mbox{FNR} = 7.4\%$,\\
$\mbox{AUCROC} = 99\%$
}
\\ \hline

$S_2$
& \makecell[l]{
$d = 10$, $\mu = 1.05$\\
$\sigma^2_A = 0.02$, $\sigma^2_B = 0.04$\\
} &
\makecell[l]{
$\mbox{FPR} = 1.0\%$,\\
$\mbox{FNR} = 15\%$,\\
$\mbox{AUCROC} = 99\%$
}
\\ \hline

\end{tabular}

\caption{Synthetic dataset simulation parameters for TvS data generation (variables are covered in Table \ref{table:dparam}) and associated ground truth metrics.}
\label{table:sim-dparam}
\ifx\templateType\templateArxiv \vspace{-4mm} \fi

\end{table} 
\subsection{Test Cases} \label{sec:scenarios}

Using the TvS distribution (Section \ref{sec:tvs}), two specific test cases were selected, named $S_1$ and $S_2$, with 2 and 10 features respectively (details of the distribution parameters are covered in Table \ref{table:dparam}).
Based on the results of \cite{wheat_testing_2025}, the TvS distribution was a very difficult case when used with 8 and more features, thus the 10 feature case ($S_2$) is also expected to be a challenge for the detectors.

Given that the simulated dataset does not have a limit to the number of examples that can be generated, Monte Carlo simulation can be used to approximate the ideal performance metrics for a distribution (Section \ref{sec:gt}).
 \pdfmarkupcomment[color=green]{In this case, 10 240 batches of 1024, for a total of 10 485 760 points for each scenario, are used to calculate the ground truth in Table {\ref{table:sim-dparam}}.}{Added.}

The distribution parameters (Table \ref{table:dparam}) were purposefully chosen to create a problem where the ideal AUCROC scores are similar and the classes are not fully separable.
This was done for two reasons: to create a more difficult problem and to more easily identify cases in which detector performance is overestimated.
While the ideal error rate is not low enough to be acceptable for many applications, using a harder problem is better for evaluating detectors.
This problem is separable enough, so that it may easily \say{appear} to be fully separable in the training dataset, even though it is not.

Remember that the overall separability of the problem is not known in practice, although it is typically assumed that the problems are \say{solvable} with the available data and a 0\% error rate is achievable.
Here, a different approach is taken, given that it is known that there will be a trade-off between the FPR and the FNR.
In this experiment, the target or desired FPR is set to 1\% and the goal is to capture as many faulty examples as possible while maintaining that FPR.
In effect, a 1\% FPR is considered to be \say{acceptable} for these problems (which may vary in practice).
It is important to note that setting the target error rate based on FPR is much more reliable than FNR due to the small number of faulty examples available for training.

\ifx\templateType\templateIEEE\else
\begin{figure}[htb]
\centering
     \includegraphics[width=0.95\linewidth]{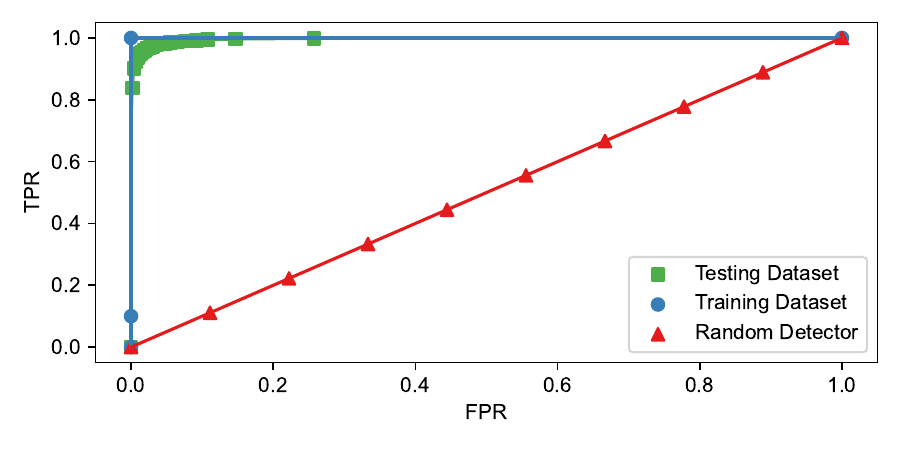}
    \caption{Example ROC curves for ideal cases in Figure \ref{fig:ex-fp-fn} and random case.}
    \label{fig:aucroc}
\end{figure} \fi
\renewcommand{\arraystretch}{1.1}
\begin{table*}[ht]
\centering
\ifx\templateType\templateIEEE
\small
\fi
\begin{tabular}{p{2cm}llll}

Category & Type & Set Hyperparameters and Value & Tuned Hyperparameters and Allowed Values \\
\hline

\parbox[t]{2mm}{\multirow{12}{*}{\rotatebox[origin=c]{90}{\makecell{One-Class\\Unsupervised (US)}}}}
&CBLOF&
None
& 
n\_clusters: 4, 6, 8, 10, 12\\ \cline{2-4}

&DeepSVDD&
None
& 
None \\ \cline{2-4}

&IForest* &
\makecell[l]{
contamination: 0.01\\
}
& 
\makecell[l]{
random\_state: 0 to 49 \\
n\_estimators: 50, 75, 100\\
max\_samples: 'auto', 0.5, 0.7, 0.8, 0.9
} \\ \cline{2-4}

&kNN &
\makecell[l]{
contamination: 0.01\\
}
& 
\ifx\templateType\templateIEEE
\makecell[l]{
n\_neighbors: 3, 5, 7, 0.001N, 0.01N, 0.02N, 0.03N, 0.04N,\\
\hspace{1.7cm} 0.06N, 0.08N, 0.10N, 0.12N, 0.15N $\dagger$
} 
\else
\makecell[l]{
n\_neighbors: 3, 5, 7, 0.001N, 0.01N, 0.02N, \\
\hspace{2cm} 0.03N, 0.04N, 0.06N, 0.08N,\\
\hspace{2cm} 0.10N, 0.12N, 0.15N $\dagger$
} 
\fi
\\\cline{2-4}

& LOF &
\makecell[l]{
contamination: 0.01\\
}
& 
\ifx\templateType\templateIEEE
\makecell[l]{
n\_neighbors: 3, 5, 7, 0.001N, 0.01N, 0.02N, 0.03N, 0.04N,\\
\hspace{1.7cm} 0.06N, 0.08N, 0.10N, 0.12N, 0.15N $\dagger$
} 
\else
\makecell[l]{
n\_neighbors: 3, 5, 7, 0.001N, 0.01N, 0.02N, \\
\hspace{2cm} 0.03N, 0.04N, 0.06N, 0.08N,\\
\hspace{2cm} 0.10N, 0.12N, 0.15N $\dagger$
} 
\fi
\\  \cline{2-4}

&OCSVM &
\makecell[l]{
contamination: 0.01
} & 
\makecell[l]{
kernel: sigmoid, rbf, linear \\
gamma: auto, scale \\
nu: 0.3, 0.5, 0.7, 0.9
} \\ 

\hline

\parbox[t]{2mm}{\multirow{4}{*}{\rotatebox[origin=c]{90}{\makecell{Two-Class\\Semi-\\supervised\\(SS)}}}}
&DeepSAD  & None & seed: 1, 2, 3, 4, 5 \\ \cline{2-4}
&DevNet & None & seed: 1, 2, 3, 4, 5  \\ \cline{2-4}
&PreNet  & None & seed: 1, 2, 3, 4, 5 \\ \cline{2-4}
&XGBOD & None 
& 
\makecell[l]{
random\_state: 1, 2, 3, 4, 5 \\
} \\ \hline

\parbox[t]{2mm}{\multirow{5}{*}{\rotatebox[origin=c]{90}{\makecell{Two-Class\\Fully\\Supervised\\(FS)}}}}
&CatBoost& None & None\\ \cline{2-4} 

&FTTransformer& None & seed: 1, 2, 3, 4, 5\\ \cline{2-4} 

&SVM &
\makecell[l]{
class\_weight: balanced \\
break\_ties: True
} & 
\makecell[l]{
kernel: sigmoid, rbf, linear \\
gamma: auto, scale \\
c: 0.1, 0.3, 0.5, 0.7, 1.0, 1.2, 1.5, 1.7, 2.0
} \\ \cline{2-4}

&XGB & None & None \\ \hline

\end{tabular}

\caption{Custom anomaly detector parameters. In cases where it is None, no hyperparameter tuning was done and default parameters were used. System specific and debug parameters (memory allocation, GPU selection, verbosity, etc.) are excluded.\\
* Indicates that only certain combinations were used. 
\ifx\templateType\templateIEEE\else \\ \fi
$\dagger$ N is the total number of training samples. Fractions are rounded up.}
\label{table:HP} \label{table:detectors}

\end{table*}
 
\subsection{Metrics} \label{sec:metrics}

As the anomaly scores for each detector are relative and can not be directly compared, the base metrics used to measure detector performance in each simulation are:

\noindent
\begin{minipage}{0.95\linewidth}
\ifx\templateType\templateArxiv\smallskip\fi
\smallskip
\begin{itemize}
    \item AUCROC: Area Under the Receiver Operating Characteristic Curve
    \item FPR: False Positive Rate
    \item FNR: False Negative Rate
\end{itemize}
\smallskip
\end{minipage}

The Receiver Operating Characteristic Curve shows the rate of true positives against false positives across all threshold settings 
\ifx\templateType\templateIEEE\else
(Figure \ref{fig:aucroc}) 
\fi
and the AUCROC is the area under the curve.
Given sets of faulty and healthy data ($X^{(f)}$ and $X^{(h)}$) with $n_f$ and $n_h$ points, respectively, the AUCROC is calculated as:
\begin{equation}
    \text{AUCROC} = \frac{\sum_{i=1}^{n_f} \sum_{j=1}^{n_h} \mathbf{I}\big( s(x_i^{(f)}) > s(x_j^{(h)}) \big)}{n_f \cdot n_h} 
\end{equation}
where $I$ is the indicator function and $s(x)$ corresponds to the raw anomaly detection score of point $x$ for each detector.
Given that the problem is already binary, no modification is needed, class imbalance is already taken into account.

AUCROC is a very common metric for evaluating anomaly detectors \cite{jourdan_machine_2021} because it condenses the trad- off between false positives and false negatives into a single metric (Figure \ref{fig:ex-fp-fn}).
However, this means that any AUCROC value may correspond to multiple different curves with different behaviors (for example, as in Table \ref{table:sim-dparam}).

Therefore, in addition, the simple anomaly predictor from Section \ref{sec:simple} is also applied to the anomaly detection scores to produce labels from the anomaly scores.
Thus, detector performance at the target FPR can be compared, as well as consistency on the test set.
While some detectors already have their own prediction functions, using a single method for this step removes inconsistencies in different implementations.

Note that since the CDF and ICDF functions used in Equations \ref{eq:fpr_to_thres} and \ref{eq:thres_to_fnr} are estimates, constructed empirically based on the available scores, their accuracy is also influenced by the number of examples.
Small sample size may cause variation in the interpolation results and  \pdfmarkupcomment[color=yellow]{jumping}{technical usage, not sure if there's a better word}.

\subsection{Detectors} \label{sec:detectors}

Based on the problem criteria (Section \ref{sec:challenges}), existing detectors were sought out that had been previously evaluated and performed well on datasets with very small numbers of anomalous examples.
Based on the results of \cite{han_adbench_2022}, the detectors were selected based on high average AUCROC scores on a large number of benchmark datasets, when only trained with 1\% of the dataset's anomalies being labelled.
Given that unlabelled anomalies still existed in the training datasets, this case is slightly different, however, the small number of labels means the problem was still difficult.

Pre-scaling is integrated into every detector (other than ground truth), so that the input is scaled to unit variance and zero mean.
The selected detectors are listed in Table \ref{table:detectors}.
Unless listed in Table \ref{table:HP}, hyperparameters were left at the default values of \cite{han_adbench_2022}.
The implementation of the detectors is kept as close as possible to the versions used in \cite{han_adbench_2022}, with some exceptions for unsupervised methods.

\ifx\templateType\templateArxiv \smallskip\fi
\subsubsection{Modifications to Unsupervised Methods}
Note that in this application, there is an assumption that the labels are available, but the purpose of this experiment is \textbf{not} to examine the benefits of using labelled vs. unlabelled data.
Thus, unsupervised detectors face a natural disadvantage (not utilizing labels) over supervised detectors.

There are known issues which can occur when giving anomalous data to the unsupervised detectors.
For example, if anomalous points start to form clusters, then kNN will start to recognize them as part of the distribution rather than anomalies.
Depending on the type of anomaly, this data may influence the detector in unexpected ways.
Thus, the unsupervised detectors are treated in this case as \say{one-class} detectors, by removing the anomalous data from all training datasets.

Due to this, and how the labels are also used in cross-validation; and inform hyperparameter selection, the \say{unsupervised} detectors are not truly unsupervised.
This terminology is simply to represent the different approaches used by the detection methods.

\subsection{Hyperparameter Selection and Validation} \label{sec:CV}

For hyperparameter selection, five-fold cross validation is used, meaning the training dataset is split into five train-test 
\ifx\templateType\templateArxiv
subsets (See Figure \ref{fig:cross-val}).
\else
subsets.
\fi
In each case, the model is trained on four of the subsets (80\%) while one is reversed for testing (20\%).
Faulty examples are evenly divided (or as closely as possible) between each subset.
Note that this requires a minimum number of five faulty examples in the original training dataset.

The validation metrics are taken as the average values over the five testing datasets.
For models with hyperparameter tuning (Table \ref{table:HP}), those values are selected based on the highest validation AUCROC.
This cross-validation step is performed even for detectors with no hyperparameter tuning to obtain validation metrics.
The validation metrics serve as predictors of model performance on the testing dataset and will be used later on for checking generalizability.

For detectors that use random initial seeds values, those seeds are included as parameters to be tuned.
For LOF and kNN, the number of neighbors used is expanded beyond the list used by \cite{han_adbench_2022} to include ratios based on the total numbers of training examples.
While the ideal hyperparameters for these detectors are known to vary, based on the number of examples \cite{fukunaga_bayes_1987}, and this change may not fully correct for that variation, the 80\% training validation size vs. a 100\% training dataset size difference, is not expected to be large enough to cause issues.

\subsection{Implementation}

The program used to conduct the simulations is provided at
{\href{https://github.com/LesleyWheat/AD-imbalanceTester}{github.com/LesleyWheat/AD-imbalanceTester}}.
For the sake of reproducibility, only anomaly detection techniques with open-source implementations were applied (Table {\ref{table:detectors}}).
Although there may be some variations from computation on different hardware (for example, CPU vs. GPU), this is expected to be negligible in the overall results.

\pdfmarkupcomment[color=green]{
All detection methods must complete successfully in order for a simulation to be counted.
During hyperparameter tuning, values which result in errors are excluded.
In the case where no valid hyperparameters are found, or the chosen hyperparameters produce an error when the method is applied to the full training dataset, the simulation is excluded from the results.
}{Added}

\ifx\templateType\templateIEEE
\begin{figure}[htb]
  \begin{minipage}{\linewidth}
\centering
    \includegraphics[width=0.95\linewidth]{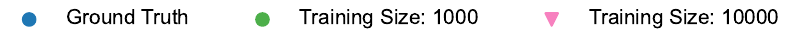}
    \subcaptionbox{Simulation test error rates of LOF on $S_1$.}
      {\includegraphics[width=0.95\linewidth]{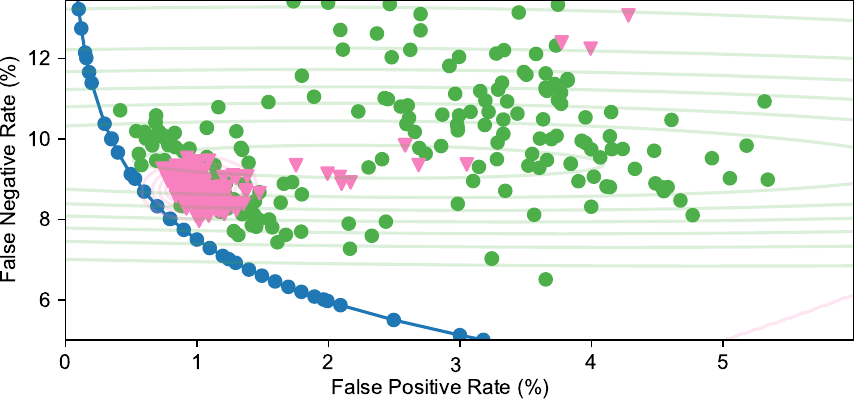}}
      \hfill
    \subcaptionbox{Simulation test error rates of XGBOD on $S_2$.}
      {\includegraphics[width=0.95\linewidth]{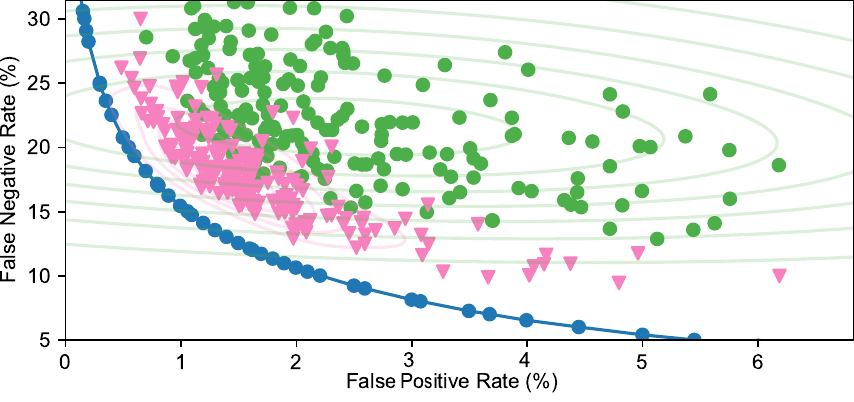}}
    \caption{Examples of detector performance differences based on the training dataset size at an anomaly rate of 0.5\%.
    Each point represents the results of a single simulation while contour plots are estimates of the distributions for each set (constructed using kernel density estimation).}
    \label{fig:fp-fn}
  \end{minipage}\quad
\end{figure} 

\ifx\templateType\templateIEEE
\begin{figure}[htb]
\else
\begin{figure}[b]
\fi
\centering
    \ifx\templateType\templateIEEE
     \includegraphics[width=0.85\linewidth]{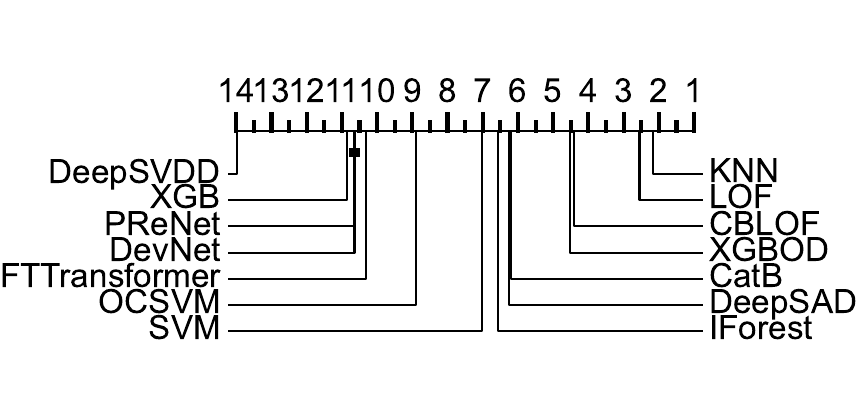}
     \else
     \includegraphics[width=0.95\linewidth]{CD-all}
     \fi
    \caption{Critical difference diagram over average rank results over all simulations with a training anomaly rate of 0.5\% (minimum 199 simulation per sample size). Results not statistically significant are connected by a black bar. Created using the same method as \cite{han_adbench_2022}. }
    \label{fig:cd}
\end{figure} 
\ifx\templateType\templateIEEE
\begin{table*}[t]
\small
\else
\begin{table*}[p]
\fi

\centering

\begin{minipage}{1\linewidth}
    \centering
    \subcaptionbox{Results for $S_1$ (2 features).
    \ifx\templateType\templateIEEE\vspace{-0.4cm}\fi
    }{\begin{tabular}{@{\extracolsep{\fill}}ccccccc}\\
&&&\multicolumn{4}{c}{Number of Training Examples}\\\cline{4-7}
&&&1000&2000&5000&10000\\ \cline{2-7}
\parbox[t]{3mm}{\multirow{15}{*}{\rotatebox[origin=c]{90}{Anomaly Detection Method}}}&True&TD&\cellcolor{gray!25}(98.9, 99.1)&(98.9, 99.1)&\cellcolor{gray!25}(98.9, 99.1)&(98.9, 99.1)\\ \cline{2-7}
&\multirow{6}{*}{US}&CBLOF&\textbf{(2.2)} \textbf{(97.1, 98.1)}&\cellcolor{gray!25}(4.4) (97.0, 98.1)&(6.7) (96.9, 98.0)&\cellcolor{gray!25}(8.2) (96.7, 98.0)\\ \cline{3-7}
&&DeepSVDD&\cellcolor{gray!25}(13.9) (34.8, 47.3)&(14.0) (39.9, 46.9)&\cellcolor{gray!25}(14.0) (42.1, 46.3)&(14.0) (44.7, 46.3)\\ \cline{3-7}
&&IForest&\textbf{(3.6)} \textbf{(96.4, 97.9)}&\cellcolor{gray!25}\textbf{(4.3)} \textbf{(97.3, 98.2)}&(5.2) (97.7, 98.2)&\cellcolor{gray!25}(5.6) (97.8, 98.2)\\ \cline{3-7}
&&KNN&\cellcolor{gray!25}\textbf{(2.4)} \textbf{(97.4, 98.4)}&\textbf{(1.7)} \textbf{(97.9, 98.5)}&\cellcolor{gray!25}\textbf{(2.3)} \textbf{(98.0, 98.5)}&\textbf{(3.0)} \textbf{(98.1, 98.5)}\\ \cline{3-7}
&&LOF&(3.9) (94.2, 98.5)&\cellcolor{gray!25}\textbf{(2.1)} \textbf{(95.8, 98.5)}&\textbf{(3.0)} \textbf{(95.6, 98.5)}&\cellcolor{gray!25}(3.3) (96.4, 98.5)\\ \cline{3-7}
&&OCSVM&\cellcolor{gray!25}(11.2) (47.9, 84.4)&(12.6) (50.5, 84.1)&\cellcolor{gray!25}(13.0) (82.8, 83.8)&(13.0) (82.8, 83.8)\\ \cline{2-7}
&\multirow{4}{*}{SS}&DeepSAD&(6.3) (92.3, 97.1)&\cellcolor{gray!25}(5.9) (94.5, 98.4)&(5.7) (95.5, 98.5)&\cellcolor{gray!25}(6.4) (96.2, 98.5)\\ \cline{3-7}
&&DevNet&\cellcolor{gray!25}(9.3) (53.0, 94.5)&(10.1) (83.8, 95.1)&\cellcolor{gray!25}(11.3) (91.2, 95.6)&(11.6) (92.9, 95.5)\\ \cline{3-7}
&&PReNet&(9.3) (52.7, 94.4)&\cellcolor{gray!25}(9.6) (81.5, 95.0)&(10.9) (92.8, 95.4)&\cellcolor{gray!25}(11.3) (94.0, 95.6)\\ \cline{3-7}
&&XGBOD&\cellcolor{gray!25}(4.4) (95.3, 98.2)&(5.5) (95.2, 98.3)&\cellcolor{gray!25}(6.1) (96.4, 98.4)&(5.9) (96.8, 98.5)\\ \cline{2-7}
&\multirow{4}{*}{FS}&CatB&(6.3) (80.2, 98.0)&\cellcolor{gray!25}(5.9) (87.3, 98.3)&(4.2) (96.8, 98.6)&\cellcolor{gray!25}\textbf{(3.0)} \textbf{(97.7, 98.7)}\\ \cline{3-7}
&&FTTransformer&\cellcolor{gray!25}(8.6) (71.5, 97.3)&(9.1) (82.8, 97.5)&\cellcolor{gray!25}(10.1) (91.4, 97.7)&(9.0) (94.0, 98.2)\\ \cline{3-7}
&&SVM&(11.9) (15.5, 94.8)&\cellcolor{gray!25}(9.0) (50.0, 98.5)&\textbf{(3.6)} \textbf{(93.6, 98.7)}&\cellcolor{gray!25}\textbf{(2.1)} \textbf{(97.6, 98.7)}\\ \cline{3-7}
&&XGB&\cellcolor{gray!25}(11.7) (51.2, 86.9)&(10.9) (76.2, 95.1)&\cellcolor{gray!25}(8.9) (93.0, 97.5)&(8.6) (96.2, 97.9)\\ \cline{2-7}
\end{tabular} }
\end{minipage}
\begin{minipage}{1\linewidth}
    \centering
    \subcaptionbox{Results for $S_2$ (10 features).}{\begin{tabular}{@{\extracolsep{\fill}}ccccccc}\\
&&&\multicolumn{4}{c}{Number of Training Examples}\\\cline{4-7}
&&&1000&2000&5000&10000\\ \cline{2-7}
\parbox[t]{3mm}{\multirow{15}{*}{\rotatebox[origin=c]{90}{Anomaly Detection Method}}}&True&TD&\cellcolor{gray!25}(99.0, 99.1)&(99.0, 99.1)&\cellcolor{gray!25}(99.0, 99.1)&(99.0, 99.1)\\ \cline{2-7}
&\multirow{6}{*}{US}&CBLOF&\textbf{(2.3)} \textbf{(95.3, 96.9)}&\cellcolor{gray!25}\textbf{(3.2)} \textbf{(95.2, 96.8)}&(4.1) (95.0, 96.7)&\cellcolor{gray!25}(4.0) (96.4, 97.6)\\ \cline{3-7}
&&DeepSVDD&\cellcolor{gray!25}(13.9) (39.7, 50.8)&(14.0) (42.1, 50.8)&\cellcolor{gray!25}(14.0) (46.5, 50.3)&(14.0) (46.7, 50.2)\\ \cline{3-7}
&&IForest&(7.1) (85.8, 93.2)&\cellcolor{gray!25}(7.5) (89.5, 94.5)&(8.7) (91.2, 94.6)&\cellcolor{gray!25}(10.7) (91.7, 95.1)\\ \cline{3-7}
&&KNN&\cellcolor{gray!25}\textbf{(1.9)} \textbf{(95.4, 97.5)}&\textbf{(1.6)} \textbf{(96.5, 97.5)}&\cellcolor{gray!25}\textbf{(1.9)} \textbf{(96.8, 97.4)}&\textbf{(2.7)} \textbf{(96.9, 97.4)}\\ \cline{3-7}
&&LOF&\textbf{(2.3)} \textbf{(94.0, 97.6)}&\cellcolor{gray!25}\textbf{(1.7)} \textbf{(95.0, 97.6)}&\textbf{(1.8)} \textbf{(96.0, 97.5)}&\cellcolor{gray!25}\textbf{(2.5)} \textbf{(96.7, 97.5)}\\ \cline{3-7}
&&OCSVM&\cellcolor{gray!25}(4.1) (94.6, 95.7)&(4.5) (94.8, 95.6)&\cellcolor{gray!25}(5.7) (95.0, 95.6)&(7.4) (95.0, 95.5)\\ \cline{2-7}
&\multirow{4}{*}{SS}&DeepSAD&(7.9) (84.5, 91.7)&\cellcolor{gray!25}(6.1) (91.9, 94.7)&(5.9) (94.0, 96.0)&\cellcolor{gray!25}(5.9) (95.2, 96.7)\\ \cline{3-7}
&&DevNet&\cellcolor{gray!25}(11.0) (65.4, 84.5)&(11.0) (74.3, 89.8)&\cellcolor{gray!25}(10.2) (88.2, 93.4)&(10.7) (91.8, 94.7)\\ \cline{3-7}
&&PReNet&(11.0) (65.5, 83.8)&\cellcolor{gray!25}(11.1) (77.9, 87.9)&(11.1) (86.5, 93.4)&\cellcolor{gray!25}(10.8) (91.5, 95.0)\\ \cline{3-7}
&&XGBOD&\cellcolor{gray!25}(5.3) (87.0, 96.3)&(4.6) (88.5, 97.1)&\cellcolor{gray!25}\textbf{(2.8)} \textbf{(94.7, 97.7)}&\textbf{(1.5)} \textbf{(96.7, 98.1)}\\ \cline{2-7}
&\multirow{4}{*}{FS}&CatB&(5.8) (88.4, 95.2)&\cellcolor{gray!25}(7.9) (88.3, 93.9)&(8.8) (90.2, 94.6)&\cellcolor{gray!25}(7.4) (93.9, 96.5)\\ \cline{3-7}
&&FTTransformer&\cellcolor{gray!25}(9.4) (68.8, 88.9)&(10.8) (70.2, 89.5)&\cellcolor{gray!25}(12.7) (78.1, 92.1)&(12.7) (85.4, 94.1)\\ \cline{3-7}
&&SVM&(10.4) (35.5, 92.2)&\cellcolor{gray!25}(8.2) (51.4, 95.1)&(6.1) (91.7, 96.7)&\cellcolor{gray!25}(4.8) (94.8, 97.5)\\ \cline{3-7}
&&XGB&\cellcolor{gray!25}(12.7) (47.7, 74.4)&(12.7) (66.0, 84.9)&\cellcolor{gray!25}(11.2) (86.4, 92.9)&(9.9) (92.3, 95.0)\\ \cline{2-7}
\end{tabular} }
\end{minipage}

\caption{
Testing bounds for AUCROC, shown in percent, for simulations with a training dataset anomaly rate of 0.5\% (minimum of 199 simulations per sample size).
The first bracket contains the average rank by test AUCROC (lower is better) (excluding true detector).
The second bracket contains the lower (2.5\%) and upper (97.5\%) bounds.
The top three detectors (excluding true detector), by average rank for each number of training samples, are in bold.
}
\label{table:results-auc-test}

\end{table*}
 \else

\ifx\templateType\templateIEEE
\begin{figure}[htb]
\else
\begin{figure}[b]
\fi
\centering
    \ifx\templateType\templateIEEE
     \includegraphics[width=0.85\linewidth]{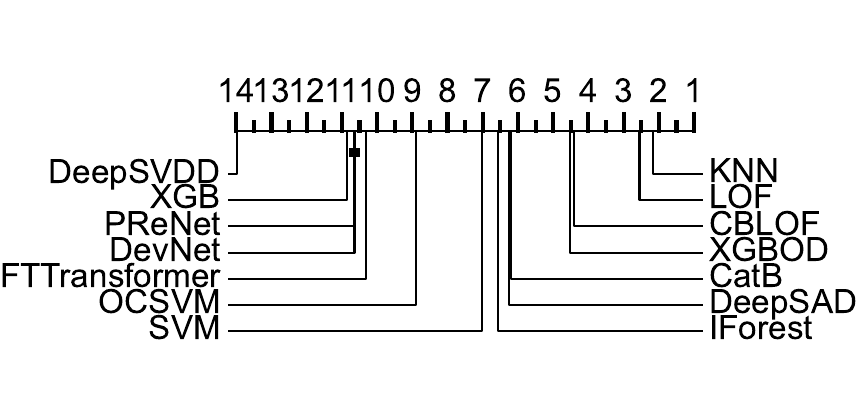}
     \else
     \includegraphics[width=0.95\linewidth]{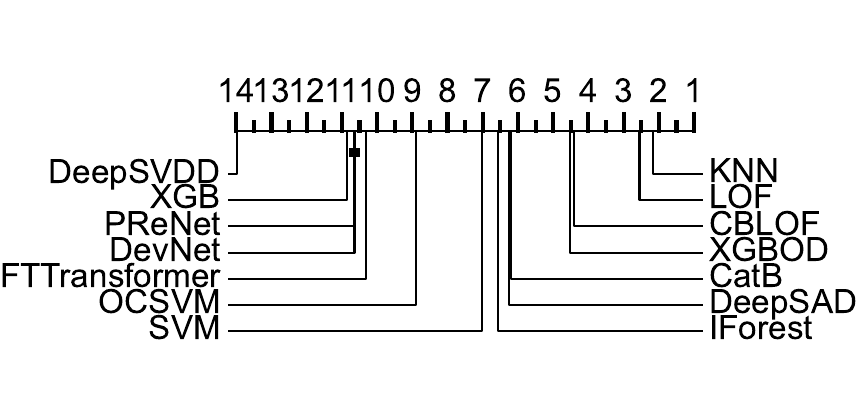}
     \fi
    \caption{Critical difference diagram over average rank results over all simulations with a training anomaly rate of 0.5\% (minimum 199 simulation per sample size). Results not statistically significant are connected by a black bar. Created using the same method as \cite{han_adbench_2022}. }
    \label{fig:cd}
\end{figure} \begin{figure}[htb]
  \begin{minipage}{\linewidth}
\centering
    \includegraphics[width=0.95\linewidth]{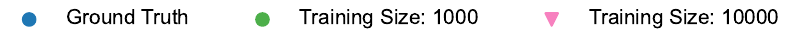}
    \subcaptionbox{Simulation test error rates of LOF on $S_1$.}
      {\includegraphics[width=0.95\linewidth]{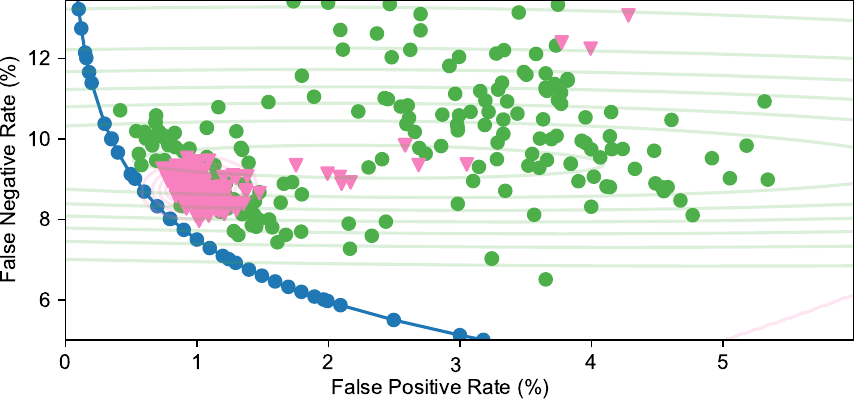}}
      \hfill
    \subcaptionbox{Simulation test error rates of XGBOD on $S_2$.}
      {\includegraphics[width=0.95\linewidth]{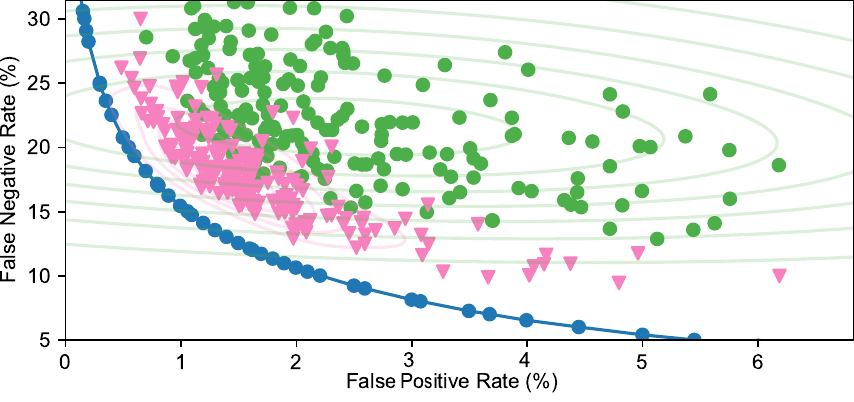}}
    \caption{Examples of detector performance differences based on the training dataset size at an anomaly rate of 0.5\%.
    Each point represents the results of a single simulation while contour plots are estimates of the distributions for each set (constructed using kernel density estimation).}
    \label{fig:fp-fn}
  \end{minipage}\quad
\end{figure} 
\ifx\templateType\templateIEEE
\begin{table*}[t]
\small
\else
\begin{table*}[p]
\fi

\centering

\begin{minipage}{1\linewidth}
    \centering
    \subcaptionbox{Results for $S_1$ (2 features).
    \ifx\templateType\templateIEEE\vspace{-0.4cm}\fi
    }{\begin{tabular}{@{\extracolsep{\fill}}ccccccc}\\
&&&\multicolumn{4}{c}{Number of Training Examples}\\\cline{4-7}
&&&1000&2000&5000&10000\\ \cline{2-7}
\parbox[t]{3mm}{\multirow{15}{*}{\rotatebox[origin=c]{90}{Anomaly Detection Method}}}&True&TD&\cellcolor{gray!25}(98.9, 99.1)&(98.9, 99.1)&\cellcolor{gray!25}(98.9, 99.1)&(98.9, 99.1)\\ \cline{2-7}
&\multirow{6}{*}{US}&CBLOF&\textbf{(2.2)} \textbf{(97.1, 98.1)}&\cellcolor{gray!25}(4.4) (97.0, 98.1)&(6.7) (96.9, 98.0)&\cellcolor{gray!25}(8.2) (96.7, 98.0)\\ \cline{3-7}
&&DeepSVDD&\cellcolor{gray!25}(13.9) (34.8, 47.3)&(14.0) (39.9, 46.9)&\cellcolor{gray!25}(14.0) (42.1, 46.3)&(14.0) (44.7, 46.3)\\ \cline{3-7}
&&IForest&\textbf{(3.6)} \textbf{(96.4, 97.9)}&\cellcolor{gray!25}\textbf{(4.3)} \textbf{(97.3, 98.2)}&(5.2) (97.7, 98.2)&\cellcolor{gray!25}(5.6) (97.8, 98.2)\\ \cline{3-7}
&&KNN&\cellcolor{gray!25}\textbf{(2.4)} \textbf{(97.4, 98.4)}&\textbf{(1.7)} \textbf{(97.9, 98.5)}&\cellcolor{gray!25}\textbf{(2.3)} \textbf{(98.0, 98.5)}&\textbf{(3.0)} \textbf{(98.1, 98.5)}\\ \cline{3-7}
&&LOF&(3.9) (94.2, 98.5)&\cellcolor{gray!25}\textbf{(2.1)} \textbf{(95.8, 98.5)}&\textbf{(3.0)} \textbf{(95.6, 98.5)}&\cellcolor{gray!25}(3.3) (96.4, 98.5)\\ \cline{3-7}
&&OCSVM&\cellcolor{gray!25}(11.2) (47.9, 84.4)&(12.6) (50.5, 84.1)&\cellcolor{gray!25}(13.0) (82.8, 83.8)&(13.0) (82.8, 83.8)\\ \cline{2-7}
&\multirow{4}{*}{SS}&DeepSAD&(6.3) (92.3, 97.1)&\cellcolor{gray!25}(5.9) (94.5, 98.4)&(5.7) (95.5, 98.5)&\cellcolor{gray!25}(6.4) (96.2, 98.5)\\ \cline{3-7}
&&DevNet&\cellcolor{gray!25}(9.3) (53.0, 94.5)&(10.1) (83.8, 95.1)&\cellcolor{gray!25}(11.3) (91.2, 95.6)&(11.6) (92.9, 95.5)\\ \cline{3-7}
&&PReNet&(9.3) (52.7, 94.4)&\cellcolor{gray!25}(9.6) (81.5, 95.0)&(10.9) (92.8, 95.4)&\cellcolor{gray!25}(11.3) (94.0, 95.6)\\ \cline{3-7}
&&XGBOD&\cellcolor{gray!25}(4.4) (95.3, 98.2)&(5.5) (95.2, 98.3)&\cellcolor{gray!25}(6.1) (96.4, 98.4)&(5.9) (96.8, 98.5)\\ \cline{2-7}
&\multirow{4}{*}{FS}&CatB&(6.3) (80.2, 98.0)&\cellcolor{gray!25}(5.9) (87.3, 98.3)&(4.2) (96.8, 98.6)&\cellcolor{gray!25}\textbf{(3.0)} \textbf{(97.7, 98.7)}\\ \cline{3-7}
&&FTTransformer&\cellcolor{gray!25}(8.6) (71.5, 97.3)&(9.1) (82.8, 97.5)&\cellcolor{gray!25}(10.1) (91.4, 97.7)&(9.0) (94.0, 98.2)\\ \cline{3-7}
&&SVM&(11.9) (15.5, 94.8)&\cellcolor{gray!25}(9.0) (50.0, 98.5)&\textbf{(3.6)} \textbf{(93.6, 98.7)}&\cellcolor{gray!25}\textbf{(2.1)} \textbf{(97.6, 98.7)}\\ \cline{3-7}
&&XGB&\cellcolor{gray!25}(11.7) (51.2, 86.9)&(10.9) (76.2, 95.1)&\cellcolor{gray!25}(8.9) (93.0, 97.5)&(8.6) (96.2, 97.9)\\ \cline{2-7}
\end{tabular} }
\end{minipage}
\begin{minipage}{1\linewidth}
    \centering
    \subcaptionbox{Results for $S_2$ (10 features).}{\begin{tabular}{@{\extracolsep{\fill}}ccccccc}\\
&&&\multicolumn{4}{c}{Number of Training Examples}\\\cline{4-7}
&&&1000&2000&5000&10000\\ \cline{2-7}
\parbox[t]{3mm}{\multirow{15}{*}{\rotatebox[origin=c]{90}{Anomaly Detection Method}}}&True&TD&\cellcolor{gray!25}(99.0, 99.1)&(99.0, 99.1)&\cellcolor{gray!25}(99.0, 99.1)&(99.0, 99.1)\\ \cline{2-7}
&\multirow{6}{*}{US}&CBLOF&\textbf{(2.3)} \textbf{(95.3, 96.9)}&\cellcolor{gray!25}\textbf{(3.2)} \textbf{(95.2, 96.8)}&(4.1) (95.0, 96.7)&\cellcolor{gray!25}(4.0) (96.4, 97.6)\\ \cline{3-7}
&&DeepSVDD&\cellcolor{gray!25}(13.9) (39.7, 50.8)&(14.0) (42.1, 50.8)&\cellcolor{gray!25}(14.0) (46.5, 50.3)&(14.0) (46.7, 50.2)\\ \cline{3-7}
&&IForest&(7.1) (85.8, 93.2)&\cellcolor{gray!25}(7.5) (89.5, 94.5)&(8.7) (91.2, 94.6)&\cellcolor{gray!25}(10.7) (91.7, 95.1)\\ \cline{3-7}
&&KNN&\cellcolor{gray!25}\textbf{(1.9)} \textbf{(95.4, 97.5)}&\textbf{(1.6)} \textbf{(96.5, 97.5)}&\cellcolor{gray!25}\textbf{(1.9)} \textbf{(96.8, 97.4)}&\textbf{(2.7)} \textbf{(96.9, 97.4)}\\ \cline{3-7}
&&LOF&\textbf{(2.3)} \textbf{(94.0, 97.6)}&\cellcolor{gray!25}\textbf{(1.7)} \textbf{(95.0, 97.6)}&\textbf{(1.8)} \textbf{(96.0, 97.5)}&\cellcolor{gray!25}\textbf{(2.5)} \textbf{(96.7, 97.5)}\\ \cline{3-7}
&&OCSVM&\cellcolor{gray!25}(4.1) (94.6, 95.7)&(4.5) (94.8, 95.6)&\cellcolor{gray!25}(5.7) (95.0, 95.6)&(7.4) (95.0, 95.5)\\ \cline{2-7}
&\multirow{4}{*}{SS}&DeepSAD&(7.9) (84.5, 91.7)&\cellcolor{gray!25}(6.1) (91.9, 94.7)&(5.9) (94.0, 96.0)&\cellcolor{gray!25}(5.9) (95.2, 96.7)\\ \cline{3-7}
&&DevNet&\cellcolor{gray!25}(11.0) (65.4, 84.5)&(11.0) (74.3, 89.8)&\cellcolor{gray!25}(10.2) (88.2, 93.4)&(10.7) (91.8, 94.7)\\ \cline{3-7}
&&PReNet&(11.0) (65.5, 83.8)&\cellcolor{gray!25}(11.1) (77.9, 87.9)&(11.1) (86.5, 93.4)&\cellcolor{gray!25}(10.8) (91.5, 95.0)\\ \cline{3-7}
&&XGBOD&\cellcolor{gray!25}(5.3) (87.0, 96.3)&(4.6) (88.5, 97.1)&\cellcolor{gray!25}\textbf{(2.8)} \textbf{(94.7, 97.7)}&\textbf{(1.5)} \textbf{(96.7, 98.1)}\\ \cline{2-7}
&\multirow{4}{*}{FS}&CatB&(5.8) (88.4, 95.2)&\cellcolor{gray!25}(7.9) (88.3, 93.9)&(8.8) (90.2, 94.6)&\cellcolor{gray!25}(7.4) (93.9, 96.5)\\ \cline{3-7}
&&FTTransformer&\cellcolor{gray!25}(9.4) (68.8, 88.9)&(10.8) (70.2, 89.5)&\cellcolor{gray!25}(12.7) (78.1, 92.1)&(12.7) (85.4, 94.1)\\ \cline{3-7}
&&SVM&(10.4) (35.5, 92.2)&\cellcolor{gray!25}(8.2) (51.4, 95.1)&(6.1) (91.7, 96.7)&\cellcolor{gray!25}(4.8) (94.8, 97.5)\\ \cline{3-7}
&&XGB&\cellcolor{gray!25}(12.7) (47.7, 74.4)&(12.7) (66.0, 84.9)&\cellcolor{gray!25}(11.2) (86.4, 92.9)&(9.9) (92.3, 95.0)\\ \cline{2-7}
\end{tabular} }
\end{minipage}

\caption{
Testing bounds for AUCROC, shown in percent, for simulations with a training dataset anomaly rate of 0.5\% (minimum of 199 simulations per sample size).
The first bracket contains the average rank by test AUCROC (lower is better) (excluding true detector).
The second bracket contains the lower (2.5\%) and upper (97.5\%) bounds.
The top three detectors (excluding true detector), by average rank for each number of training samples, are in bold.
}
\label{table:results-auc-test}

\end{table*}
 \fi

\section{Results}

\subsection{Test Performance} \label{sec:anomalyRate} \label{sec:results-test}

The overall ranks from the simulation testing AUCROC results are reported in Figure \ref{fig:cd} to compare against the results in \cite{han_adbench_2022}.
In the critical difference diagram, the datasets sizes are all combined, and show kNN and LOF as the top detectors.
However, this picture does not tell the entire story, and from the expanded results in Table \ref{table:results-auc-test}, the ranking of the detectors is highly dependent on the total dataset size.

Looking at the performance for different training dataset sizes, the top detectors on the $S_1$ case are kNN, LOF and SVM.
For the smaller training datasets, kNN is the best detector but as the size increases, SVM performance increases.
In $S_2$, kNN is also the best detector for small datasets but XGBOD ranks at the top for larger datasets.

Most detectors had decreased performance on $S_2$ compared to $S_1$, as expected due to the increased difficultly of the problem.
Additionally, the AUCROC ranges tend to also be larger, reflecting increased performance variability.
Interestingly, OCSVM does perform better with more features.

Figure \ref{fig:fp-fn} displays the raw results for LOF and XGBOD, illustrating how detectors approach the performance of the ground truth detector with more examples; but tend to have variation in both the false positive and false negative rate.
It can also be noted, that the results tend to follow a similar curve to the ground truth, as expected.
This also indicates that the false positive and negative rates do not follow a Gaussian distribution.

\ifx\templateType\templateIEEE
\begin{figure}[t]
\else
\begin{figure}[tb]
\fi
  \begin{minipage}{\linewidth}
\centering
    
\includegraphics[width=0.95\linewidth]{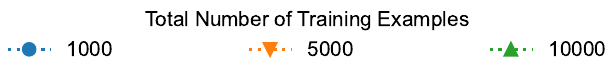}

    \subcaptionbox{$S_1$: Unsupervised.}
      {\includegraphics[width=0.48\linewidth]{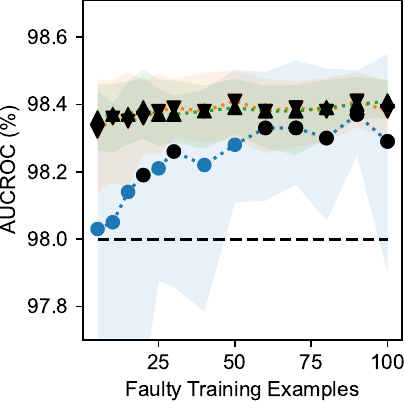}}
      \hfill
    \subcaptionbox{$S_2$: Unsupervised.}
      {\includegraphics[width=0.48\linewidth]{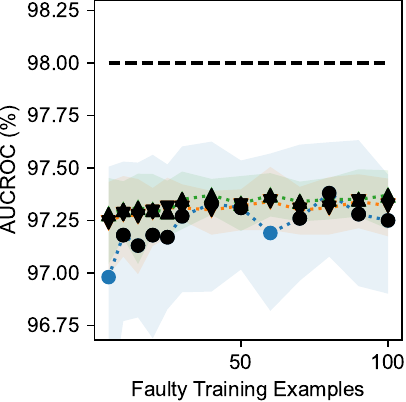}}

    \subcaptionbox{$S_1$: Semi-supervised.}
      {\includegraphics[width=0.48\linewidth]{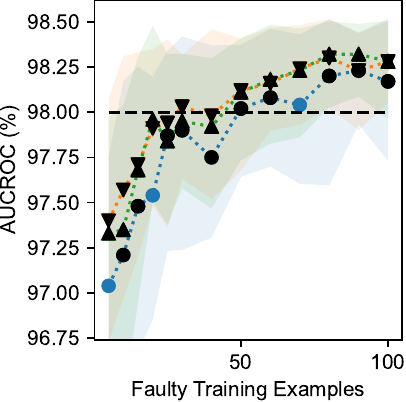}}
      \hfill
    \subcaptionbox{$S_2$: Semi-supervised.}
      {\includegraphics[width=0.48\linewidth]{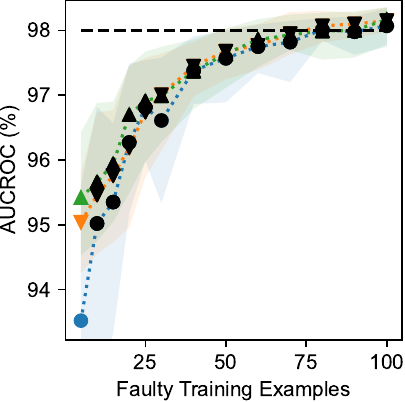}}

    \subcaptionbox{$S_1$: Supervised.}
      {\includegraphics[width=0.48\linewidth]{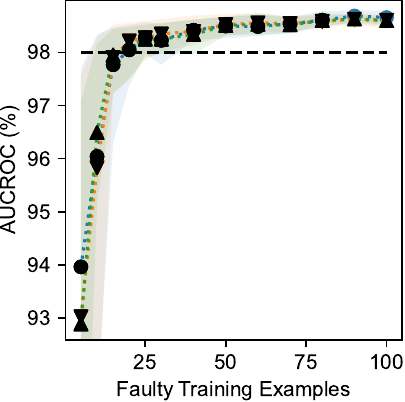}}
      \hfill
    \subcaptionbox{$S_2$: Supervised.}
      {\includegraphics[width=0.48\linewidth]{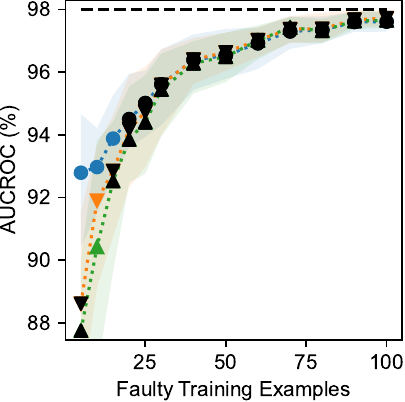}}
      
    \caption{
    AUCROC by category, for differing numbers of anomalous samples in the training dataset and the training dataset size.
Filled areas represent the 80\% range (10\% to 90\% percentile).
    Icons represent the average for each set of simulations (minimum of 25 simulations) and black icons indicate results that are not statistically significantly different from all other dataset sizes ($p > 0.05$).
    The black dashed line is at 98\% AUCROC for reference.
    }
    \label{fig:count-vs-rate}
  \end{minipage}\quad
\end{figure} 
\ifx\templateType\templateIEEE
\begin{table}[t]
\else
\begin{table}[tb]
\fi

\centering
\ifx\templateType\templateIEEE
\small
\fi

\begin{minipage}{0.9\linewidth}
    \centering
    \subcaptionbox{Results for $S_1$ (2 features).
    \ifx\templateType\templateIEEE\vspace{-0.3cm}\fi
    }{\begin{tabular}{@{\extracolsep{\fill}}ccccc}\\
&&\multicolumn{3}{c}{\makecell{Total Number of\\Training Examples}}\\\cline{3-5}
&&1000&5000&10000\\ \cline{2-5}
\parbox[t]{8mm}{\multirow{11}{*}{\rotatebox[origin=c]{90}{\makecell{Number of Anomaly\\Training Examples}}}}&5&\cellcolor{gray!25}US&US&\cellcolor{gray!25}US\\ \cline{2-5}
&10&US&\cellcolor{gray!25}US&US\\ \cline{2-5}
&15&\cellcolor{gray!25}\textbf{US}&US&\cellcolor{gray!25}US\\ \cline{2-5}
&20&\textbf{US}&\cellcolor{gray!25}\textbf{US}&US\\ \cline{2-5}
&25&\cellcolor{gray!25}\textbf{FS}&\textbf{US}&\cellcolor{gray!25}\textbf{US}\\ \cline{2-5}
&30&\textbf{US}&\cellcolor{gray!25}\textbf{US}&\textbf{US}\\ \cline{2-5}
&40&\cellcolor{gray!25}FS&\textbf{FS}&\cellcolor{gray!25}\textbf{US}\\ \cline{2-5}
&50&FS&\cellcolor{gray!25}FS&FS\\ \cline{2-5}
&60&\cellcolor{gray!25}FS&FS&\cellcolor{gray!25}FS\\ \cline{2-5}
&70&FS&\cellcolor{gray!25}FS&FS\\ \cline{2-5}
&80&\cellcolor{gray!25}FS&FS&\cellcolor{gray!25}FS\\ \cline{2-5}
\end{tabular} }
\end{minipage}
\begin{minipage}{0.9\linewidth}
    \centering
    \subcaptionbox{Results for $S_2$ (10 features).}{\begin{tabular}{@{\extracolsep{\fill}}ccccc}\\
&&\multicolumn{3}{c}{\makecell{Total Number of\\Training Examples}}\\\cline{3-5}
&&1000&5000&10000\\ \cline{2-5}
\parbox[t]{8mm}{\multirow{11}{*}{\rotatebox[origin=c]{90}{\makecell{Number of Anomaly\\Training Examples}}}}&5&\cellcolor{gray!25}US&US&\cellcolor{gray!25}US\\ \cline{2-5}
&10&US&\cellcolor{gray!25}US&US\\ \cline{2-5}
&15&\cellcolor{gray!25}US&US&\cellcolor{gray!25}US\\ \cline{2-5}
&20&US&\cellcolor{gray!25}US&US\\ \cline{2-5}
&25&\cellcolor{gray!25}US&US&\cellcolor{gray!25}US\\ \cline{2-5}
&30&US&\cellcolor{gray!25}\textbf{US}&US\\ \cline{2-5}
&40&\cellcolor{gray!25}\textbf{SS}&SS&\cellcolor{gray!25}\textbf{SS}\\ \cline{2-5}
&50&SS&\cellcolor{gray!25}SS&SS\\ \cline{2-5}
&60&\cellcolor{gray!25}SS&SS&\cellcolor{gray!25}SS\\ \cline{2-5}
&70&SS&\cellcolor{gray!25}SS&SS\\ \cline{2-5}
&80&\cellcolor{gray!25}SS&SS&\cellcolor{gray!25}SS\\ \cline{2-5}
\end{tabular} }
\end{minipage}

\caption{
Best detection strategy by average maximum AUCROC by number of anomaly examples and total number of training examples (minimum of 25 simulations).
Categories from Table \ref{table:detectors}.
Bold indicates the results are not statistically significantly different from all other methods ($p > 0.05$).
}
\label{table:results-categories}

\end{table} 

\begin{figure*}[t]
  \begin{minipage}{\linewidth}
\centering
  \includegraphics[width=0.3\linewidth]{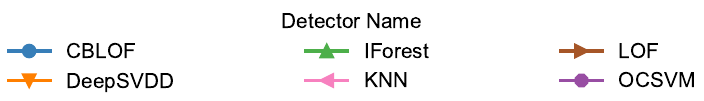}
  \hfill
    \includegraphics[width=0.3\linewidth]{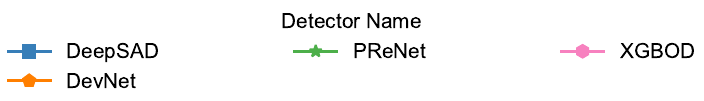}
    \hfill
      \includegraphics[width=0.3\linewidth]{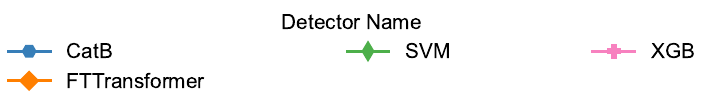}
    \subcaptionbox{Unsupervised AUC MSE.}
      {\includegraphics[width=0.3\linewidth]{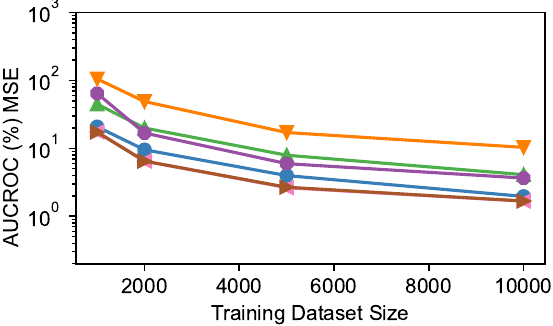}}
      \hfill
    \subcaptionbox{Semi-supervised AUC MSE.}
      {\includegraphics[width=0.3\linewidth]{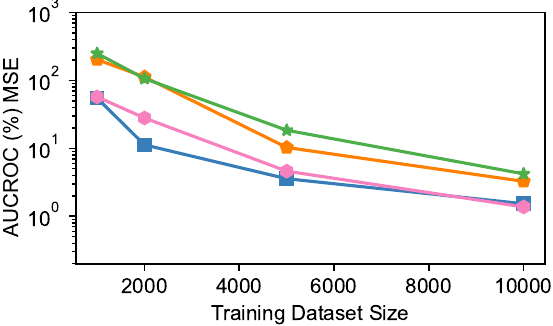}}
      \hfill
    \subcaptionbox{Supervised AUC MSE.}
      {\includegraphics[width=0.3\linewidth]{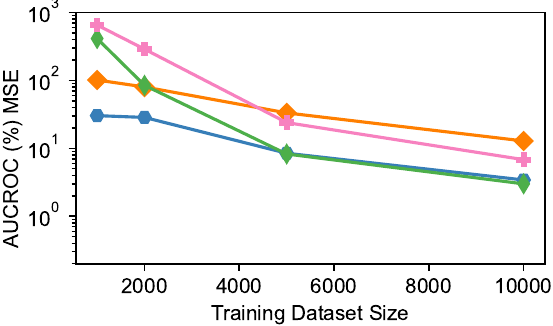}}
    \subcaptionbox{Unsupervised FNR MSE.}
      {\includegraphics[width=0.3\linewidth]{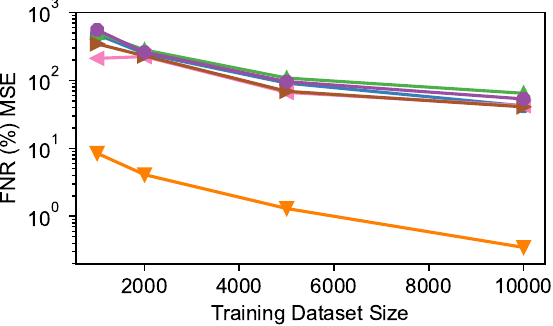}}
      \hfill
    \subcaptionbox{Semi-supervised FNR MSE.}
      {\includegraphics[width=0.3\linewidth]{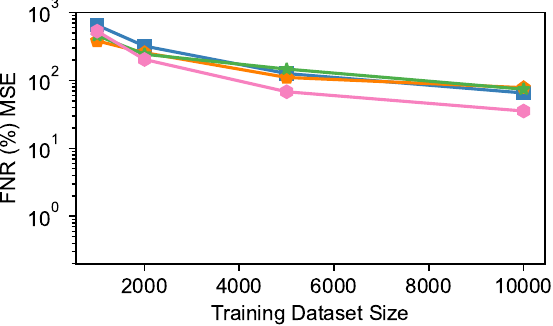}}
      \hfill
    \subcaptionbox{Supervised FNR MSE.}
      {\includegraphics[width=0.3\linewidth]{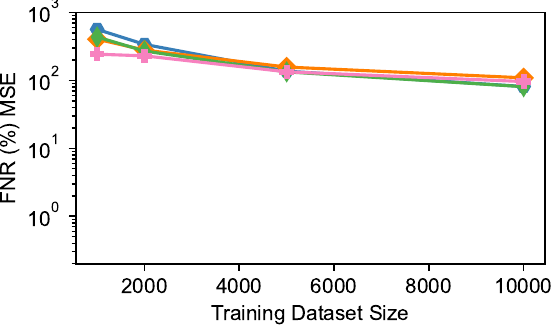}}

    \caption{Generalization results of $S_2$ with a training anomaly rate of 0.5\% for different training dataset sizes (minimum of 199 simulations per sample size).}
    \label{fig:auc-vs-fnr}
  \end{minipage}\quad
\end{figure*} 
To look more broadly at the performance over the different types of approaches, in both cases, unsupervised methods comprise the top detectors on the smaller datasets (Table \ref{table:results-auc-test}).
As more examples are added, the performance of the top unsupervised methods do not increase significantly, but methods which utilize labels quickly improve as more examples are added.
The change in rank is not due to a decrease in unsupervised detector performance, but rather an increase in semi-supervised and supervised detector performance.
As well, in the larger feature space ($S_2$), semi-supervised methods do tend to perform better than the supervised methods.

From the results in  Table \ref{table:results-auc-test}, it is not possible to know if the change in detector performance is due to the increase in faulty examples, or in healthy examples in the training dataset.
To examine the impact of adding more of each type of class to the training dataset, more simulations are conducted with differing anomaly rates, to create the results in Figure \ref{fig:count-vs-rate}.
To simplify the information, the detectors are grouped into categories (as detailed in Table {\ref{table:detectors}}) and, for each simulation, the maximum AUCROC, over all detectors in that category, is taken as the category metric.

With 50 simulations per combination of dataset size and number of faulty examples, the results across different numbers of healthy examples are rarely statistically significant (at $p \leq 0.05$).
Most statistically significant results are from the smallest dataset (1 000 examples) and where, as expected, the unsupervised methods struggle.
Overall, the impact of additional healthy examples to the training dataset is low.

The curves in Figure \ref{fig:count-vs-rate} do have the same general shape as the results reported by \cite{han_adbench_2022}.
While the results of \cite{han_adbench_2022} show most semi-supervised algorithms outperform the best unsupervised methods at 1\% labelled anomaly rate, these results show that behaviour only occurs when a minimum amount of labelled faulty examples are available.
In \cite{han_adbench_2022}, the 1\% labelled anomaly rate corresponds to different numbers of total labelled anomalies in the training dataset, due to different dataset sizes and base anomaly rates.
Here, it is shown that the change in performance is primarily due to the total number of faulty examples in training.
These curves are extended on the left side, showing where supervised methods significantly suffer and unsupervised methods have the highest performance.

Based on these results, the best detection strategy is largely determined by the total number of anomaly examples available for training.
This is further explored by Table {\ref{table:results-categories}}, which shows the detection category with the highest average maximum AUCROC for each dataset makeup.
From this table, it is clear that unless there are at least 20-50 anomaly examples available for training, unsupervised methods are the best option.

While the transition from unsupervised occurs at about the same number of faulty examples for $S_1$ and $S_2$, that is where the similarities end.
For $S_1$, it is better to transition directly to a supervised approach rather than using semi-supervised learning.
But in $S_2$, it is the opposite, where semi-supervised methods perform better than supervised in all tested cases.
Thus, it is apparent that semi-supervised methods have an advantage over supervised methods in larger feature spaces.

\ifx\templateType\templateIEEE
\begin{table*}[t]
\small
\else
\begin{table*}[p]
\fi

\centering

\begin{minipage}{1\linewidth}
    \centering
    \subcaptionbox{Results for $S_1$ (2 features).
    \ifx\templateType\templateIEEE\vspace{-0.3cm}\fi
    }{\begin{tabular}{@{\extracolsep{\fill}}ccccccc}\\
&&&\multicolumn{4}{c}{Number of Training Examples}\\\cline{4-7}
&&&1000&2000&5000&10000\\ \cline{2-7}
\parbox[t]{3mm}{\multirow{15}{*}{\rotatebox[origin=c]{90}{Anomaly Detection Method}}}&True&TD&\cellcolor{gray!25}(-1.1, 4.7)&(-1.1, 4.8)&\cellcolor{gray!25}(-1.0, 2.6)&(-1.0, 1.9)\\ \cline{2-7}
&\multirow{6}{*}{US}&CBLOF&\textbf{(3.6)} \textbf{(-2.9, 10.0)}&\cellcolor{gray!25}(6.1) (-2.8, 6.1)&(7.6) (-2.8, 3.6)&\cellcolor{gray!25}(8.2) (-2.5, 2.6)\\ \cline{3-7}
&&DeepSVDD&\cellcolor{gray!25}(7.8) (-15.5, 12.5)&(8.6) (-13.1, 9.6)&\cellcolor{gray!25}(9.0) (-6.9, 6.6)&(10.1) (-4.9, 4.3)\\ \cline{3-7}
&&IForest&\textbf{(3.6)} \textbf{(-3.0, 7.1)}&\cellcolor{gray!25}\textbf{(4.7)} \textbf{(-2.3, 6.0)}&(6.3) (-1.9, 4.2)&\cellcolor{gray!25}(6.8) (-1.7, 2.5)\\ \cline{3-7}
&&KNN&\cellcolor{gray!25}\textbf{(3.9)} \textbf{(-2.6, 6.7)}&\textbf{(3.9)} \textbf{(-1.8, 6.9)}&\cellcolor{gray!25}\textbf{(5.6)} \textbf{(-1.7, 3.7)}&(6.3) (-1.5, 2.5)\\ \cline{3-7}
&&LOF&(5.1) (-4.7, 6.3)&\cellcolor{gray!25}\textbf{(4.1)} \textbf{(-2.8, 5.3)}&(5.9) (-3.7, 3.1)&\cellcolor{gray!25}\textbf{(6.2)} \textbf{(-2.2, 2.2)}\\ \cline{3-7}
&&OCSVM&\cellcolor{gray!25}(10.0) (-48.2, 16.8)&(10.2) (-25.2, 13.8)&\cellcolor{gray!25}(11.2) (-10.6, 10.7)&(11.0) (-7.1, 9.4)\\ \cline{2-7}
&\multirow{4}{*}{SS}&DeepSAD&(6.2) (-7.5, 8.0)&\cellcolor{gray!25}(6.4) (-4.6, 4.5)&(6.4) (-2.7, 4.1)&\cellcolor{gray!25}(6.5) (-2.2, 2.1)\\ \cline{3-7}
&&DevNet&\cellcolor{gray!25}(9.4) (-45.2, 33.0)&(8.3) (-8.1, 13.8)&\cellcolor{gray!25}(8.1) (-4.3, 4.3)&(7.8) (-2.8, 3.4)\\ \cline{3-7}
&&PReNet&(9.9) (-44.0, 32.6)&\cellcolor{gray!25}(8.6) (-9.7, 14.6)&(8.2) (-4.3, 5.9)&\cellcolor{gray!25}(7.9) (-2.8, 3.5)\\ \cline{3-7}
&&XGBOD&\cellcolor{gray!25}(6.0) (-4.4, 11.4)&(7.9) (-4.7, 8.9)&\cellcolor{gray!25}(7.2) (-3.5, 3.8)&(7.0) (-2.4, 2.6)\\ \cline{2-7}
&\multirow{4}{*}{FS}&CatB&(7.8) (-19.1, 17.2)&\cellcolor{gray!25}(6.4) (-9.6, 10.5)&\textbf{(5.4)} \textbf{(-1.8, 4.1)}&\cellcolor{gray!25}\textbf{(5.3)} \textbf{(-1.3, 2.6)}\\ \cline{3-7}
&&FTTransformer&\cellcolor{gray!25}(8.7) (-26.1, 20.6)&(9.1) (-13.1, 10.4)&\cellcolor{gray!25}(9.7) (-7.0, 4.2)&(8.7) (-4.1, 3.8)\\ \cline{3-7}
&&SVM&(11.7) (-54.7, 21.8)&\cellcolor{gray!25}(10.5) (-37.2, 17.6)&\textbf{(5.6)} \textbf{(-3.3, 8.0)}&\cellcolor{gray!25}\textbf{(4.7)} \textbf{(-1.4, 2.2)}\\ \cline{3-7}
&&XGB&\cellcolor{gray!25}(11.1) (-39.0, 55.0)&(10.2) (-17.0, 25.7)&\cellcolor{gray!25}(8.6) (-4.1, 8.1)&(8.5) (-2.4, 4.2)\\ \cline{2-7}
\end{tabular} }
\end{minipage}
\begin{minipage}{1\linewidth}
    \centering
    \subcaptionbox{Results for $S_2$ (10 features).}{\begin{tabular}{@{\extracolsep{\fill}}ccccccc}\\
&&&\multicolumn{4}{c}{Number of Training Examples}\\\cline{4-7}
&&&1000&2000&5000&10000\\ \cline{2-7}
\parbox[t]{3mm}{\multirow{15}{*}{\rotatebox[origin=c]{90}{Anomaly Detection Method}}}&True&TD&\cellcolor{gray!25}(-1.0, 5.8)&(-1.0, 2.9)&\cellcolor{gray!25}(-0.8, 1.5)&(-0.7, 1.0)\\ \cline{2-7}
&\multirow{6}{*}{US}&CBLOF&\textbf{(4.2)} \textbf{(-4.3, 12.5)}&\cellcolor{gray!25}\textbf{(5.3)} \textbf{(-4.0, 7.3)}&(6.3) (-3.3, 3.6)&\cellcolor{gray!25}(6.4) (-2.1, 3.0)\\ \cline{3-7}
&&DeepSVDD&\cellcolor{gray!25}(8.3) (-22.8, 11.0)&(8.8) (-15.2, 8.6)&\cellcolor{gray!25}(9.4) (-9.6, 5.5)&(9.8) (-7.3, 4.6)\\ \cline{3-7}
&&IForest&(7.5) (-12.3, 10.9)&\cellcolor{gray!25}(7.6) (-7.9, 5.9)&(8.2) (-5.5, 4.0)&\cellcolor{gray!25}(8.6) (-3.8, 4.0)\\ \cline{3-7}
&&KNN&\cellcolor{gray!25}\textbf{(4.0)} \textbf{(-4.4, 12.2)}&\textbf{(3.9)} \textbf{(-3.0, 6.8)}&\cellcolor{gray!25}\textbf{(5.0)} \textbf{(-2.4, 3.6)}&\textbf{(5.8)} \textbf{(-1.8, 3.0)}\\ \cline{3-7}
&&LOF&\textbf{(3.9)} \textbf{(-5.5, 9.5)}&\cellcolor{gray!25}\textbf{(3.7)} \textbf{(-2.9, 7.3)}&\textbf{(5.0)} \textbf{(-2.3, 3.6)}&\cellcolor{gray!25}\textbf{(5.8)} \textbf{(-1.8, 3.1)}\\ \cline{3-7}
&&OCSVM&\cellcolor{gray!25}(5.9) (-5.0, 16.2)&(6.4) (-4.6, 9.6)&\cellcolor{gray!25}(6.9) (-3.6, 5.5)&(8.1) (-2.6, 5.6)\\ \cline{2-7}
&\multirow{4}{*}{SS}&DeepSAD&(8.0) (-13.2, 5.5)&\cellcolor{gray!25}(6.0) (-6.1, 3.7)&\textbf{(6.2)} \textbf{(-3.3, 1.6)}&\cellcolor{gray!25}(6.3) (-2.4, 2.2)\\ \cline{3-7}
&&DevNet&\cellcolor{gray!25}(10.1) (-25.6, 27.0)&(9.7) (-18.1, 21.1)&\cellcolor{gray!25}(8.0) (-6.1, 6.0)&(7.5) (-3.5, 3.5)\\ \cline{3-7}
&&PReNet&(10.4) (-29.6, 29.1)&\cellcolor{gray!25}(10.0) (-17.9, 22.8)&(9.2) (-8.1, 8.6)&\cellcolor{gray!25}(8.2) (-4.4, 3.9)\\ \cline{3-7}
&&XGBOD&\cellcolor{gray!25}(7.5) (-12.0, 15.4)&(7.0) (-11.1, 10.3)&\cellcolor{gray!25}(6.6) (-4.1, 4.2)&\textbf{(5.4)} \textbf{(-2.1, 2.3)}\\ \cline{2-7}
&\multirow{4}{*}{FS}&CatB&(5.9) (-8.0, 12.2)&\cellcolor{gray!25}(7.5) (-7.5, 11.2)&(7.5) (-4.0, 6.8)&\cellcolor{gray!25}(7.1) (-2.4, 4.8)\\ \cline{3-7}
&&FTTransformer&\cellcolor{gray!25}(8.2) (-22.8, 12.0)&(9.2) (-20.8, 9.6)&\cellcolor{gray!25}(10.0) (-14.2, 5.2)&(10.4) (-7.5, 4.1)\\ \cline{3-7}
&&SVM&(10.2) (-46.9, 18.6)&\cellcolor{gray!25}(8.3) (-20.3, 18.0)&(7.2) (-3.9, 6.8)&\cellcolor{gray!25}(7.0) (-2.1, 4.1)\\ \cline{3-7}
&&XGB&\cellcolor{gray!25}(10.9) (-31.4, 59.0)&(11.6) (-22.4, 35.4)&\cellcolor{gray!25}(9.5) (-5.9, 12.2)&(8.7) (-3.7, 5.8)\\ \cline{2-7}
\end{tabular} }
\end{minipage}
\caption{
Using validation AUCROC to predict testing AUCROC with a training anomaly rate is 0.5\% (minimum of 199 simulations per sample size).
The first bracket contains the average rank by lowest squared error (excluding true detector).
The second bracket contains the lower (2.5\%) and upper (97.5\%) bounds of the prediction error.
The most consistent three detectors for each number of samples by average rank are in bold.
}
\label{table:results-auc-gen}

\end{table*}
 \ifx\templateType\templateIEEE
\begin{table*}[htb]
\small
\else
\begin{table*}[tb]
\fi

\centering

\begin{minipage}{1\linewidth}
    \centering
    \subcaptionbox{Results for $S_1$ (2 features).
    \ifx\templateType\templateIEEE\vspace{-0.3cm}\fi
    }{\begin{tabular}{@{\extracolsep{\fill}}cw{c}{0.17\textwidth}w{c}{0.17\textwidth}w{c}{0.17\textwidth}w{c}{0.17\textwidth}}\\
&\multicolumn{4}{c}{Number of Training Examples}\\\cline{2-5}
&1000&2000&5000&10000\\ \cline{1-5}
\makecell[l]{Top Five\\Detectors\\(Validation)}&\cellcolor{gray!25}\makecell[c]{XGBOD (28\%)\\ LOF (22\%)\\ KNN (13\%)\\ CBLOF (11\%)\\ FTTransformer (8\%)}&\makecell[c]{LOF (33\%)\\ CBLOF (17\%)\\ DeepSAD (16\%)\\ KNN (15\%)\\ XGBOD (9\%)}&\cellcolor{gray!25}\makecell[c]{LOF (27\%)\\ KNN (20\%)\\ SVM (13\%)\\ CBLOF (11\%)\\ XGBOD (9\%)}&\makecell[c]{SVM (22\%)\\ KNN (18\%)\\ LOF (14\%)\\ CatB (13\%)\\ XGBOD (9\%)}\\ \cline{1-5}
\makecell[l]{Top Five\\Detectors\\(Test)}&\makecell[c]{CBLOF (37\%)\\ LOF (31\%)\\ KNN (19\%)\\ IForest (6\%)\\ CatB (6\%)}&\cellcolor{gray!25}\makecell[c]{LOF (45\%)\\ KNN (42\%)\\ SVM (7\%)\\ DeepSAD (3\%)\\ XGBOD (1\%)}&\makecell[c]{SVM (33\%)\\ KNN (26\%)\\ LOF (24\%)\\ CatB (13\%)\\ DeepSAD (3\%)}&\cellcolor{gray!25}\makecell[c]{SVM (57\%)\\ CatB (25\%)\\ LOF (8\%)\\ KNN (6\%)\\ XGBOD (3\%)}\\ \cline{1-5}
Prediction Bounds&\cellcolor{gray!25}(-47.6, 1.6)&(-12.6, 1.5)&\cellcolor{gray!25}(-5.2, 1.5)&(-3.7, 1.3)\\ \cline{1-5}
\end{tabular} }
\end{minipage}
\begin{minipage}{1\linewidth}
    \centering
    \subcaptionbox{Results for $S_2$ (10 features).}{\begin{tabular}{@{\extracolsep{\fill}}cw{c}{0.17\textwidth}w{c}{0.17\textwidth}w{c}{0.17\textwidth}w{c}{0.17\textwidth}}\\
&\multicolumn{4}{c}{Number of Training Examples}\\\cline{2-5}
&1000&2000&5000&10000\\ \cline{1-5}
\makecell[l]{Top Five\\Detectors\\(Validation)}&\cellcolor{gray!25}\makecell[c]{LOF (25\%)\\ KNN (22\%)\\ CBLOF (15\%)\\ XGBOD (14\%)\\ DeepSAD (9\%)}&\makecell[c]{LOF (36\%)\\ KNN (29\%)\\ XGBOD (12\%)\\ CBLOF (11\%)\\ DeepSAD (5\%)}&\cellcolor{gray!25}\makecell[c]{XGBOD (37\%)\\ LOF (26\%)\\ KNN (21\%)\\ DeepSAD (13\%)\\ CBLOF (3\%)}&\makecell[c]{XGBOD (54\%)\\ LOF (19\%)\\ KNN (9\%)\\ DeepSAD (8\%)\\ CBLOF (7\%)}\\ \cline{1-5}
\makecell[l]{Top Five\\Detectors\\(Test)}&\makecell[c]{LOF (42\%)\\ KNN (35\%)\\ CBLOF (22\%)\\ XGBOD (0\%)}&\cellcolor{gray!25}\makecell[c]{LOF (53\%)\\ KNN (44\%)\\ XGBOD (3\%)}&\makecell[c]{LOF (48\%)\\ KNN (26\%)\\ XGBOD (26\%)\\ CBLOF (0\%)}&\cellcolor{gray!25}\makecell[c]{XGBOD (80\%)\\ LOF (9\%)\\ KNN (7\%)\\ CBLOF (3\%)\\ SVM (2\%)}\\ \cline{1-5}
Prediction Bounds&\cellcolor{gray!25}(-13.1, 1.5)&(-9.4, 2.4)&\cellcolor{gray!25}(-4.1, 1.5)&(-2.2, 2.1)\\ \cline{1-5}
\end{tabular} }
\end{minipage}

\caption{
Top five most frequently selected top detectors by validation and test AUCROC across different training dataset sizes (minimum of 199 simulations per sample size, thus 0\% may be shown due to rounding).
Percent frequency of detector selection is in brackets, rounded to nearest percent.
Training anomaly rate is 0.5\%.
}
\label{table:results-gen-rank}

\end{table*}
 
\subsection{Generalization}

Given that the testing dataset would not be available in a real-world case for detector evaluation, conclusions about detector performance would be drawn from the validation metrics.
To examine the consistency and reliability of the detector results, this section will examine the differences in results between the validation and testing metrics.

Firstly, it is worthwhile to look at changes across different performance metrics.
Figure \ref{fig:auc-vs-fnr} compares the MSE of AUCROC and FNR across the different detection methods.
In this case, the validation metric can be seen as a predictor variable for the test result.
When both measures are taken in percentage, AUCROC is much more consistent for most detectors.
Therefore, AUCROC will be used for comparison purposes.

Table {\ref{table:results-auc-gen}} shows the error range when using the validation AUCROC to predict testing AUCROC.
\pdfmarkupcomment[color=green]{
As an example, kNN in $S_1$ at 10 000 samples has an expected variation of $(-1.5, 2.5)$.
Therefore, if the validation AUCROC of kNN was measured at 97\%, the testing AUCROC would be expected to be between 95.5\% and 99.5\%.
In this way, the difference between the validation and testing metrics can be compared, with a smaller range corresponding to a more consistent detector.
}{Added example.}
Based on the results in Table {\ref{table:results-auc-gen}}, dataset size and feature size have a large impact on detector consistency.

As would be expected, the highest ranked detectors from Table {\ref{table:results-auc-test}} are generally the most consistent detectors, and are also ranked highly in Table {\ref{table:results-auc-gen}}.
However, the top average rankings in Table {\ref{table:results-auc-gen}} are not as high as the top average ranks in Table {\ref{table:results-auc-test}}, indicating that many of the detectors are performing similarly on generalization.
Broadly, the unsupervised detectors tend to be the more consistent overall, simply because they do better on the smaller datasets.

Note that detectors do not always perform worse on the testing set than the training set, and the error can be quite symmetric for some methods.
While this means detector performance has the potential to increase on the testing dataset (rather than only dropping), both drops and jumps in performance pose potential problems due to the requirement to select a detector based only on the validation \pdfmarkupcomment[color=yellow]{metrics (as only the training dataset would be available in application).}{Edited}
A higher estimated performance on validation means there is a greater chance than a poorer detector is selected for use, and a lower estimated performance may mean that the best detector is not selected.
Either case may lead to incorrect conclusions about the best detector.

In order to quantify this risk, including detector selection by validation, the best detector can be selected by validation AUCROC, whose performance can then be compared to the testing dataset.
The prediction bounds for this strategy are reported in Table \ref{table:results-gen-rank}.
These errors tend to be unsymmetrical, showing a greater tendency for AUCROC to drop between validation and testing than most detectors on their own (Table \ref{table:results-auc-gen}).

Looking at the top detectors for each set, the best testing detectors are often the best selected based on validation, as would be expected, but there are some deviations.
In several cases, deep learning methods such as DeepSAD and FTTransformer show up in validation, but not testing.
Additionally, XGBOD only shows up as a top detector in $S_2$; it shows up in the validation selection for $S_1$ as well.
This indicates that detection methods that utilize a large number of parameters are more likely to be incorrectly selected as the best detectors when they are not, which is likely contributing to the performance drop.
This may be caused by the detectors overfitting to the validation dataset.

On the other side, SVM frequently shows up as a top detector for $S_1$ at 10 000 training examples, but is not in the most frequent detectors selected by validation.
kNN and LOF both tend to be frequently selected as top performers.

In all cases, the error bounds for the prediction is very asymmetrical, with a large tendency to overestimate detection performance.
So, while many individual detectors have the chance to increase or decrease in performance, the process of selecting the detector frequently leads to a performance drop.
This could possibly be due to the cross-validation setup but the general process of applying a decision on which model to use, is itself part of the model creation process, and may contribute to over fitting.

However, the only way to avoid such a decision would be to have advance knowledge of which detector to use, or more information about the types of anomalies in the dataset.
Increased knowledge of the problem may help in narrowing down the number of possible successful detection methods; lowering the risk of inadvertently selecting a poor detector.
It can be noted that the performance of the detectors tends to become more consistent (Table \ref{table:results-auc-gen}) and closer together (Table \ref{table:results-auc-test}) as the dataset size increases, which helps to decrease the performance drop even if the best detector is not selected.

\section{Conclusion}

In this simulation study, several anomaly detection algorithms were applied to two synthetic distributions under a wide range of training conditions.
In this way, the experiment could be modified in ways that typical benchmark datasets could not be, in order to test the performance and generalization of the anomaly detectors.
These results offer further insights into detector behaviour seen in previous research, and show why future researchers should carefully consider how the makeup (number of features/number of examples/anomaly rate) of the datasets they use for benchmarking may influence their final results.

To summarize the key findings:

\begin{itemize}
    \item Only a small  number of examples (30-50 examples) are required for top semi-supervised/supervised methods to match the performance of top unsupervised methods.
    \item At only 10 features, semi-supervised methods show significant increases in performance over supervised methods for the purpose of imbalanced classification.
    \item The total number of available faulty training examples is a key factor in selecting the best anomaly detector and more important than the anomaly rate.
    \item The false negative rate has a much higher MSE on generalization than AUCROC, and caution should be taken when evaluating model performance based on validation FNR.
\end{itemize}

By simulating realistic constraints and evaluating detectors across multiple training dataset sizes, these results provide practical guidance for deploying anomaly detection in industrial environments.
They also underscore the importance of considering generalization error when selecting models for real-world applications, a characteristic that can't be tested in practice before a system is fully implemented.
Future work may include the use of model-based simulations to create further testing datasets for specific problems, or the incorporation of more anomaly detection methods.
 
\ifx\templateType\templateArxiv
\section{Acknowledgments}
This work was supported in part by the Canada Research Chairs (CRC) Program under Project CRC-2020-0127 and in part by the Natural Sciences and Engineering Research Council of Canada (NSERC) Ford-Mitacs Alliance under Project ALLRP-590906-23.
 The authors would like to kindly thank 
Joanne Doucet
for their valuable feedback on drafts of this article.

 \vspace{20mm}
\else
\fi

\ifx\templateType\templateArxiv
    \bibliographystyle{IEEEtran.bst}
\else \ifx\templateType\templateIEEE
    \bibliographystyle{IEEEtran.bst}
\else \ifx\templateType\templateElsevier
    \bibliographystyle{elsarticle-num-names}
    \fi
\fi
\fi

\FloatBarrier
\bibliography{bib/pub.bib}

\ifx\templateType\templateArxiv
    \appendices
\else \ifx\templateType\templateIEEE
    \begin{IEEEbiography}[{\includegraphics[width=1in,height=1.25in,clip,keepaspectratio]{a1.png}}]{Lesley Wheat} received the B. Eng. in mechatronics engineering from McMaster University in 2020. Currently, she is a graduate student pursuing her Ph.D. in software engineering at McMaster University. Her research interests include fault detection and diagnosis, machine learning, data complexity measures and embedded sensor systems.

\end{IEEEbiography}

\begin{IEEEbiography}[{\includegraphics[width=1in,height=1.25in,clip,keepaspectratio]{a2.png}}]{Martin v. Mohrenschildt} received the Ph.D. degree in mathematics from the ETH-Zürich, Zurich, Switzerland, in 1994. He is a Faculty Member with McMaster University, Hamilton, ON, Canada, initially in Electrical and Computer Engineering and then in Computing and Software. From 2005 to 2011, he was the Chair of the Department. During this time, he implemented the Mechatronics Program with McMaster University. While his core expertise is in mathematics, Dr. Mohrenschildt always had a big passion for electrical and mechanical systems. Over the years, Dr. Mohrenschildt and his students built several software/hardware systems for industrial applications that at their core use signal processing-based feature extraction. For the mining industry he continues to develop a system of sensors that are used commercially to analyze vibrating screens. He also designed and built a minivan sized fully immersive 6 DOF flight/driving simulator to study the effect of motion cuing in learning. His research interests include signal processing and control, sensors and data acquisition, and real time data processing.
\end{IEEEbiography}

\begin{IEEEbiography}[{\includegraphics[width=1in,height=1.25in,clip,keepaspectratio]{a3.png}}]{SAEID HABIBI} (Member, IEEE) received the Ph.D. degree in control engineering from the University of Cambridge, U.K., in 1990. He was the Chair of the Department of Mechanical Engineering, from 2008 to 2013. He is currently the Tier I Canada Research Chair and a Full Professor with the Department of Mechanical Engineering, McMaster University. He is the Founder and the Director of the Centre for Mechatronics and Hybrid Technologies (CMHT). CHMT has created one of the most advanced and best equipped automotive research laboratories in Canada and internationally specializing in predictive algorithms, tracking, prognostics, and diagnostics. CMHT currently supports a number of large projects and being a resource to automotive companies and start-ups. CMHT research is supported by industry and funding from Ontario Research Fund Research Excellence Awards and NSERC. He has published over 200 peerreviewed articles. His academic background includes research into artificial intelligence, intelligent control, state and parameter estimation, fault diagnosis, and prediction. He is a fellow of American Society of Mechanical Engineers (ASME) and Canadian Society of Mechanical Engineers (CSME).
\end{IEEEbiography}

     \if\templateSubType\templateIEEEAccess
    \EOD \appendices
    \fi
\else \ifx\templateType\templateElsevier
\fi
\fi
\fi

\ifx\templateType\templateArxiv

\FloatBarrier
\clearpage
\section{Additional Results} \label{sec:app-results}

\begin{table}[tbh]
\centering

\begin{minipage}{1\linewidth}
    \centering
    \subcaptionbox{FPR by detector for $S_1$.}{\begin{tabular}{@{\extracolsep{\fill}}ccccccc}\\
&&&\multicolumn{4}{c}{Number of Training Examples}\\\cline{4-7}
&&&1000&2000&5000&10000\\ \cline{2-7}
\parbox[t]{3mm}{\multirow{15}{*}{\rotatebox[origin=c]{90}{Anomaly Detection Method}}}&True&TD&\cellcolor{gray!25}(0.5, 1.7)&(0.6, 1.5)&\cellcolor{gray!25}(0.7, 1.3)&(0.8, 1.2)\\ \cline{2-7}
&\multirow{6}{*}{US}&CBLOF&(0.5, 2.0)&\cellcolor{gray!25}(0.6, 1.6)&(0.8, 1.4)&\cellcolor{gray!25}(0.8, 1.2)\\ \cline{3-7}
&&DeepSVDD&\cellcolor{gray!25}(0.5, 1.8)&(0.6, 1.6)&\cellcolor{gray!25}(0.8, 1.4)&(0.8, 1.2)\\ \cline{3-7}
&&IForest&(0.4, 1.5)&\cellcolor{gray!25}(0.6, 1.4)&(0.8, 1.4)&\cellcolor{gray!25}(0.9, 1.4)\\ \cline{3-7}
&&KNN&\cellcolor{gray!25}(0.9, 100.0)&(0.6, 2.6)&\cellcolor{gray!25}(0.8, 1.7)&(0.8, 1.6)\\ \cline{3-7}
&&LOF&(0.6, 71.6)&\cellcolor{gray!25}(0.6, 4.4)&(0.8, 3.8)&\cellcolor{gray!25}(0.8, 2.6)\\ \cline{3-7}
&&OCSVM&\cellcolor{gray!25}(0.5, 1.8)&(0.6, 1.4)&\cellcolor{gray!25}(0.7, 1.3)&(0.8, 1.3)\\ \cline{2-7}
&\multirow{4}{*}{SS}&DeepSAD&(0.5, 1.9)&\cellcolor{gray!25}(0.7, 1.5)&(0.8, 1.4)&\cellcolor{gray!25}(0.8, 1.2)\\ \cline{3-7}
&&DevNet&\cellcolor{gray!25}(0.5, 1.6)&(0.6, 1.4)&\cellcolor{gray!25}(0.7, 1.3)&(0.8, 1.2)\\ \cline{3-7}
&&PReNet&(0.5, 1.7)&\cellcolor{gray!25}(0.6, 1.4)&(0.8, 1.3)&\cellcolor{gray!25}(0.8, 1.2)\\ \cline{3-7}
&&XGBOD&\cellcolor{gray!25}(0.7, 8.0)&(0.8, 5.5)&\cellcolor{gray!25}(0.5, 5.2)&(0.3, 3.6)\\ \cline{2-7}
&\multirow{4}{*}{FS}&CatB&(0.7, 2.3)&\cellcolor{gray!25}(0.7, 1.8)&(0.9, 1.5)&\cellcolor{gray!25}(0.9, 1.4)\\ \cline{3-7}
&&FTTransformer&\cellcolor{gray!25}(0.5, 1.9)&(0.6, 1.6)&\cellcolor{gray!25}(0.7, 1.3)&(0.8, 1.2)\\ \cline{3-7}
&&SVM&(0.5, 1.7)&\cellcolor{gray!25}(0.6, 1.4)&(0.7, 1.4)&\cellcolor{gray!25}(0.8, 1.2)\\ \cline{3-7}
&&XGB&\cellcolor{gray!25}(0.6, 3.7)&(0.7, 1.8)&\cellcolor{gray!25}(0.8, 1.5)&(0.9, 1.4)\\ \cline{2-7}
\end{tabular} }
\end{minipage}

\begin{minipage}{1\linewidth}
    \centering
    \subcaptionbox{FNR by detector for $S_1$.}{\begin{tabular}{@{\extracolsep{\fill}}ccccccc}\\
&&&\multicolumn{4}{c}{Number of Training Examples}\\\cline{4-7}
&&&1000&2000&5000&10000\\ \cline{2-7}
\parbox[t]{3mm}{\multirow{15}{*}{\rotatebox[origin=c]{90}{Anomaly Detection Method}}}&True&TD&\cellcolor{gray!25}(6.3, 9.1)&(6.6, 8.8)&\cellcolor{gray!25}(6.8, 8.3)&(7.0, 8.0)\\ \cline{2-7}
&\multirow{6}{*}{US}&CBLOF&(8.7, 14.5)&\cellcolor{gray!25}(9.2, 14.5)&(9.4, 14.5)&\cellcolor{gray!25}(9.7, 15.3)\\ \cline{3-7}
&&DeepSVDD&\cellcolor{gray!25}(98.9, 99.9)&(98.9, 99.9)&\cellcolor{gray!25}(99.2, 99.9)&(99.8, 99.9)\\ \cline{3-7}
&&IForest&(23.1, 56.8)&\cellcolor{gray!25}(12.5, 41.1)&(9.2, 19.4)&\cellcolor{gray!25}(8.9, 10.9)\\ \cline{3-7}
&&KNN&\cellcolor{gray!25}(0.0, 9.3)&(7.2, 10.0)&\cellcolor{gray!25}(7.8, 9.5)&(8.0, 9.3)\\ \cline{3-7}
&&LOF&(6.7, 13.4)&\cellcolor{gray!25}(7.8, 11.4)&(8.1, 11.0)&\cellcolor{gray!25}(8.2, 9.5)\\ \cline{3-7}
&&OCSVM&\cellcolor{gray!25}(37.1, 99.9)&(37.8, 64.8)&\cellcolor{gray!25}(38.3, 40.7)&(38.4, 40.6)\\ \cline{2-7}
&\multirow{4}{*}{SS}&DeepSAD&(15.2, 33.4)&\cellcolor{gray!25}(11.6, 29.5)&(11.0, 27.3)&\cellcolor{gray!25}(11.5, 25.5)\\ \cline{3-7}
&&DevNet&\cellcolor{gray!25}(28.6, 74.1)&(28.1, 62.6)&\cellcolor{gray!25}(28.0, 43.0)&(27.9, 35.2)\\ \cline{3-7}
&&PReNet&(29.3, 69.6)&\cellcolor{gray!25}(28.2, 63.1)&(27.8, 39.4)&\cellcolor{gray!25}(28.2, 33.7)\\ \cline{3-7}
&&XGBOD&\cellcolor{gray!25}(7.2, 18.3)&(6.8, 13.7)&\cellcolor{gray!25}(6.6, 14.0)&(6.8, 13.2)\\ \cline{2-7}
&\multirow{4}{*}{FS}&CatB&(17.8, 61.7)&\cellcolor{gray!25}(12.5, 49.0)&(11.3, 25.9)&\cellcolor{gray!25}(10.5, 18.4)\\ \cline{3-7}
&&FTTransformer&\cellcolor{gray!25}(15.3, 69.0)&(13.6, 47.3)&\cellcolor{gray!25}(12.2, 30.3)&(10.0, 19.4)\\ \cline{3-7}
&&SVM&(31.8, 100.0)&\cellcolor{gray!25}(11.2, 77.6)&(9.3, 26.6)&\cellcolor{gray!25}(8.9, 18.1)\\ \cline{3-7}
&&XGB&\cellcolor{gray!25}(43.1, 85.9)&(33.7, 64.5)&\cellcolor{gray!25}(17.2, 38.4)&(13.0, 26.9)\\ \cline{2-7}
\end{tabular} }
\end{minipage}

\caption{
Results from simple anomaly predictor for $S_1$.
}
\label{table:results-s1}

\end{table}
 \clearpage
\begin{table}[tbh]
\centering

\begin{minipage}{1\linewidth}
    \centering
    \subcaptionbox{FPR by detector for $S_2$.}{\begin{tabular}{@{\extracolsep{\fill}}ccccccc}\\
&&&\multicolumn{4}{c}{Number of Training Examples}\\\cline{4-7}
&&&1000&2000&5000&10000\\ \cline{2-7}
\parbox[t]{3mm}{\multirow{15}{*}{\rotatebox[origin=c]{90}{Anomaly Detection Method}}}&True&TD&\cellcolor{gray!25}(0.5, 1.6)&(0.7, 1.5)&\cellcolor{gray!25}(0.7, 1.4)&(0.8, 1.2)\\ \cline{2-7}
&\multirow{6}{*}{US}&CBLOF&(0.6, 2.2)&\cellcolor{gray!25}(0.6, 1.6)&(0.7, 1.4)&\cellcolor{gray!25}(0.8, 1.3)\\ \cline{3-7}
&&DeepSVDD&\cellcolor{gray!25}(0.5, 1.7)&(0.6, 1.5)&\cellcolor{gray!25}(0.7, 1.3)&(0.8, 1.2)\\ \cline{3-7}
&&IForest&(0.4, 1.8)&\cellcolor{gray!25}(0.6, 1.5)&(0.7, 1.3)&\cellcolor{gray!25}(0.8, 1.2)\\ \cline{3-7}
&&KNN&\cellcolor{gray!25}(0.6, 100.0)&(0.6, 2.2)&\cellcolor{gray!25}(0.7, 1.8)&(0.8, 1.7)\\ \cline{3-7}
&&LOF&(0.6, 6.8)&\cellcolor{gray!25}(0.6, 4.8)&(0.7, 2.4)&\cellcolor{gray!25}(0.8, 1.5)\\ \cline{3-7}
&&OCSVM&\cellcolor{gray!25}(0.6, 2.0)&(0.6, 1.6)&\cellcolor{gray!25}(0.7, 1.4)&(0.8, 1.3)\\ \cline{2-7}
&\multirow{4}{*}{SS}&DeepSAD&(0.9, 2.6)&\cellcolor{gray!25}(0.8, 2.0)&(0.8, 1.5)&\cellcolor{gray!25}(0.9, 1.3)\\ \cline{3-7}
&&DevNet&\cellcolor{gray!25}(0.5, 1.9)&(0.6, 1.6)&\cellcolor{gray!25}(0.7, 1.3)&(0.8, 1.2)\\ \cline{3-7}
&&PReNet&(0.6, 1.9)&\cellcolor{gray!25}(0.7, 1.6)&(0.8, 1.4)&\cellcolor{gray!25}(0.8, 1.3)\\ \cline{3-7}
&&XGBOD&\cellcolor{gray!25}(1.0, 9.7)&(0.9, 6.3)&\cellcolor{gray!25}(0.7, 4.2)&(0.7, 4.1)\\ \cline{2-7}
&\multirow{4}{*}{FS}&CatB&(1.3, 3.1)&\cellcolor{gray!25}(1.2, 2.5)&(1.2, 2.1)&\cellcolor{gray!25}(1.2, 1.8)\\ \cline{3-7}
&&FTTransformer&\cellcolor{gray!25}(0.9, 2.5)&(1.0, 2.2)&\cellcolor{gray!25}(1.0, 1.9)&(1.1, 1.7)\\ \cline{3-7}
&&SVM&(0.5, 3.6)&\cellcolor{gray!25}(0.6, 1.9)&(0.7, 1.4)&\cellcolor{gray!25}(0.8, 1.2)\\ \cline{3-7}
&&XGB&\cellcolor{gray!25}(0.8, 2.1)&(0.9, 2.0)&\cellcolor{gray!25}(1.1, 1.8)&(1.2, 1.7)\\ \cline{2-7}
\end{tabular} }
\end{minipage}

\begin{minipage}{1\linewidth}
    \centering
    \subcaptionbox{FNR by detector for $S_2$.}{\begin{tabular}{@{\extracolsep{\fill}}ccccccc}\\
&&&\multicolumn{4}{c}{Number of Training Examples}\\\cline{4-7}
&&&1000&2000&5000&10000\\ \cline{2-7}
\parbox[t]{3mm}{\multirow{15}{*}{\rotatebox[origin=c]{90}{Anomaly Detection Method}}}&True&TD&\cellcolor{gray!25}(11.8, 20.1)&(12.6, 18.5)&\cellcolor{gray!25}(13.0, 17.6)&(13.8, 17.1)\\ \cline{2-7}
&\multirow{6}{*}{US}&CBLOF&(22.0, 35.9)&\cellcolor{gray!25}(24.3, 36.6)&(25.6, 36.7)&\cellcolor{gray!25}(21.2, 29.1)\\ \cline{3-7}
&&DeepSVDD&\cellcolor{gray!25}(98.8, 100.0)&(98.9, 100.0)&\cellcolor{gray!25}(98.9, 100.0)&(99.0, 100.0)\\ \cline{3-7}
&&IForest&(53.2, 83.9)&\cellcolor{gray!25}(50.3, 73.4)&(46.9, 63.2)&\cellcolor{gray!25}(43.6, 60.8)\\ \cline{3-7}
&&KNN&\cellcolor{gray!25}(0.0, 29.3)&(19.7, 28.3)&\cellcolor{gray!25}(21.2, 26.8)&(21.6, 25.9)\\ \cline{3-7}
&&LOF&(16.3, 30.6)&\cellcolor{gray!25}(19.6, 28.9)&(21.4, 27.0)&\cellcolor{gray!25}(21.7, 26.4)\\ \cline{3-7}
&&OCSVM&\cellcolor{gray!25}(27.4, 40.8)&(29.6, 39.1)&\cellcolor{gray!25}(31.1, 37.0)&(31.7, 36.1)\\ \cline{2-7}
&\multirow{4}{*}{SS}&DeepSAD&(48.4, 67.5)&\cellcolor{gray!25}(44.5, 63.4)&(41.3, 60.3)&\cellcolor{gray!25}(35.5, 51.0)\\ \cline{3-7}
&&DevNet&\cellcolor{gray!25}(70.6, 88.6)&(66.4, 83.2)&\cellcolor{gray!25}(58.3, 73.3)&(51.9, 66.8)\\ \cline{3-7}
&&PReNet&(67.0, 87.1)&\cellcolor{gray!25}(63.4, 80.7)&(54.1, 71.6)&\cellcolor{gray!25}(49.1, 64.7)\\ \cline{3-7}
&&XGBOD&\cellcolor{gray!25}(14.3, 43.2)&(12.6, 30.4)&\cellcolor{gray!25}(11.6, 28.2)&(10.7, 25.1)\\ \cline{2-7}
&\multirow{4}{*}{FS}&CatB&(35.6, 70.5)&\cellcolor{gray!25}(43.3, 65.2)&(43.7, 62.7)&\cellcolor{gray!25}(35.3, 49.8)\\ \cline{3-7}
&&FTTransformer&\cellcolor{gray!25}(72.6, 89.4)&(65.7, 87.1)&\cellcolor{gray!25}(56.6, 82.3)&(48.5, 72.7)\\ \cline{3-7}
&&SVM&(34.9, 98.9)&\cellcolor{gray!25}(37.3, 91.2)&(35.4, 67.3)&\cellcolor{gray!25}(29.7, 52.0)\\ \cline{3-7}
&&XGB&\cellcolor{gray!25}(81.7, 93.0)&(72.3, 84.3)&\cellcolor{gray!25}(53.0, 68.8)&(40.8, 52.6)\\ \cline{2-7}
\end{tabular} }
\end{minipage}

\caption{
Results from simple anomaly predictor for $S_2$.
}
\label{table:results-s2}

\end{table} 
 \fi

\end{document}